\documentstyle[leqno,titlepage]{article}

\makeatletter

\columnwidth \textwidth

\def\marginnote#1{
}

\marginnote{To Appear: Specifying Syntactic Structures}

\newcounter{subequation}[equation]
\def\thesubequation{\theequation\@alph\c@subequation}
\def\@subeqnnum{{\rm (\thesubequation)}}
\def\slabel#1{\@bsphack\if@filesw {\let\thepage\relax
   \xdef\@gtempa{\write\@auxout{\string
      \newlabel{#1}{{\thesubequation}{\thepage}}}}}\@gtempa
   \if@nobreak \ifvmode\nobreak\fi\fi\fi\@esphack}
\def\subeqnarray{\stepcounter{equation}
\let\@currentlabel=\theequation\global\c@subequation\@ne
\global\@eqnswtrue
\global\@eqcnt\z@\tabskip\@centering\let\\=\@subeqncr
$$\halign to \displaywidth\bgroup\@eqnsel\hskip\@centering
  $\displaystyle\tabskip\z@{##}$&\global\@eqcnt\@ne
  \hskip 2\arraycolsep \hfil${##}$\hfil
  &\global\@eqcnt\tw@ \hskip 2\arraycolsep $\displaystyle\tabskip\z@{##}$\hfil
   \tabskip\@centering&\llap{##}\tabskip\z@\cr}

\def\endsubeqnarray{\@@subeqncr\egroup
                     $$\global\@ignoretrue}

\def\@subeqncr{{\ifnum0=`}\fi\@ifstar{\global\@eqpen\@M
    \@ysubeqncr}{\global\@eqpen\interdisplaylinepenalty \@ysubeqncr}}

\def\@ysubeqncr{\@ifnextchar [{\@xsubeqncr}{\@xsubeqncr[\z@]}}

\def\@xsubeqncr[#1]{\ifnum0=`{\fi}\@@subeqncr
   \noalign{\penalty\@eqpen\vskip\jot\vskip #1\relax}}

\def\@@subeqncr{\let\@tempa\relax
    \ifcase\@eqcnt \def\@tempa{& & &}\or \def\@tempa{& &}
      \else \def\@tempa{&}\fi
     \@tempa \if@eqnsw\@subeqnnum\refstepcounter{subequation}\fi
     \global\@eqnswtrue\global\@eqcnt\z@\cr}

\let\@ssubeqncr=\@subeqncr
\@namedef{subeqnarray*}{\def\@subeqncr{\nonumber\@ssubeqncr}\subeqnarray}
\@namedef{endsubeqnarray*}{\global\advance\c@equation\m@ne%
                           \nonumber\endsubeqnarray}

\newcounter{enums}

\newdimen\widelabel
\def\enumsentence{\@ifnextchar[{\@enumsentence}
{\refstepcounter{enums}\@enumsentence[(\theenums)]}}

\long\def\@enumsentence[#1]#2{\begin{list}{}{%
\advance\leftmargin by\widelabel \advance\labelwidth by \widelabel}
\item[#1] #2
\end{list}}

\newcounter{tempcnt}

\def\@item[#1]{\if@noparitem \@donoparitem
  \else \if@inlabel \indent \par \fi
         \ifhmode \unskip\unskip \par \fi 
         \if@newlist \if@nobreak \@nbitem \else
                        \addpenalty\@beginparpenalty
                        \addvspace\@topsep \addvspace{-\parskip}\fi
           \else \addpenalty\@itempenalty \addvspace\itemsep 
          \fi 
    \global\@inlabeltrue 
\fi
\everypar{\global\@minipagefalse\global\@newlistfalse 
          \if@inlabel\global\@inlabelfalse \hskip -\parindent \box\@labels
             \penalty\z@ \fi
          \everypar{}}\global\@nobreakfalse
\if@noitemarg \@noitemargfalse \if@nmbrlist \refstepcounter{\@listctr}\fi \fi
\setbox\@tempboxa\hbox{\makelabel{#1}}%
\global\setbox\@labels
 \hbox{\unhbox\@labels \hskip \itemindent
       \hskip -\labelwidth \hskip -\labelsep 
       \ifdim \wd\@tempboxa >\labelwidth 
                \box\@tempboxa
          \else \hbox to\labelwidth {\unhbox\@tempboxa}\fi
       \hskip \labelsep}\ignorespaces}

\newcounter{enumsi}

\newdimen\eeindent
\def\@mklab#1{\hfil#1}
\def\enummklab#1{\hfil(\eelabel)\hbox to \eeindent{\hfil#1}}
\def\enummakelabel#1{\enummklab{#1}\global\let\makelabel=\@mklab}
\def\toplabel#1{{\edef\@currentlabel{\p@enums\theenums}\label{#1}}}

\def\eenumsentence{\@ifnextchar[{\@eenumsentence}
{\refstepcounter{enums}\@eenumsentence[\theenums]}}

\long\def\@eenumsentence[#1]#2{\def\eelabel{#1}\let\holdlabel\makelabel%
\begin{list}{\alph{enumsi}.}{\usecounter{enumsi}%
\advance\leftmargin by \eeindent \advance\leftmargin by \widelabel%
\advance\labelwidth by \eeindent \advance\labelwidth by \widelabel%
\let\makelabel=\enummakelabel}
#2
\end{list}\let\makelabel\holdlabel}

\def\Nat{{\Bbb N}}

\font\tenmsb=msbm10
\font\sevenmsb=msbm7
\font\fivemsb=msbm5
\newfam\msbfam
 \textfont\msbfam=\tenmsb  \scriptfont\msbfam=\sevenmsb
  \scriptscriptfont\msbfam=\fivemsb

\def\Bbb{\ifmmode\let\next\Bbb@\else
 \def\next{\errmessage{Use \string\Bbb\space only in math mode}}\fi\next}
\def\Bbb@#1{{\Bbb@@{#1}}}
\def\Bbb@@#1{\fam\msbfam#1}
\newcommand{\set}[1]{\left\{
#1\right\}}

\newcommand{\bm}[1]{\mbox{\boldmath $#1$}}
\newcommand{\bms}[1]{{\mbox{\boldmath \scriptsize $#1$}}}

\newcommand{\f}[1]{{\mathchoice{\mathop{\mbox{\rm #1}}}%
{\mathop{\mbox{\rm #1}}}{\mathop{\mbox{\scriptsize\rm #1}}}%
{\mathop{\mbox{\scriptsize\rm #1}}}}}

\newcommand{\xp}[1]{%
\ifmmode\mathchoice{\mathord{\mbox{#1P}}}{\mathord{\mbox{#1P}}}%
{\mathord{\mbox{\scriptsize#1P}}}{\mathord{\mbox{\scriptsize#1P}}}%
\else\mbox{#1P}%
\fi}

\newcommand{\xb}[1]{%
\ifmmode\mathchoice{\overline{\rm #1}}{\overline{\rm #1}}%
{\mbox{\scriptsize$\overline{\rm #1}$}}{\mbox{\scriptsize$\overline{\rm #1}$}}%
\else
\mbox{\rlap{$\overline{\rm #1}$}\phantom{\rm #1}}%
\fi}

\newcommand{\xz}[1]{%
\ifmmode\mathchoice{\mathord{\rm #1}^0}{\mathord{\rm #1}^0}%
{\mathord{\scriptsize\rm #1}^0}{\mathord{\scriptsize\rm #1}^0}%
\else
\mbox{${\rm #1}^0$}%
\fi}

\newcommand{\sub}[1]{\ifmmode{_{#1}}\else$_{#1}$\fi}

\newcommand{\Abar}{\mbox{\rlap{$\overline{\rm A}$}\phantom{\rm A}}}
\newcommand{\Rbar}{\mbox{\rlap{$\overline{\rm Ref}$}\phantom{\rm Ref}}}

\newtheorem{definition}{Definition}
\newtheorem{corollary}{Corollary}
\newtheorem{theorem}{Theorem}

\def\parent{\mathrel{\triangleleft}}
\def\prdom{\mathrel{\triangleleft^+}}
\def\dom{\mathrel{\triangleleft^*}}
\def\eq{\mathrel{\approx}}
\def\lft{\mathrel{\prec}}

\let\limp=\rightarrow

\newcommand{\tup}[1]%
  {\left\langle
#1\right\rangle}

\newcommand{\sU}{{\cal U}}
\newcommand{\sI}{{\cal I}}
\newcommand{\sP}{{\cal P}}
\newcommand{\sL}{{\cal L}}
\newcommand{\sR}{{\cal R}}
\newcommand{\sD}{{\cal D}}
\newcommand{\sN}{{\cal N}}
\newcommand{\sT}{{\cal T}}

\def\IMP{\mathrel{\Rightarrow}}
\def\IFF{\mathrel{\Leftrightarrow}}
\let\liff=\leftrightarrow
\let\lif=\leftarrow
\newcommand{\tiff}{\mbox{ iff }}

\newcommand{\pair}[2]%
  {\left\langle
#1,#2\right\rangle}

\def\Th{\mathop{\bf Th}\nolimits}
\def\Mod{\mathop{\bf Mod}\nolimits}

\def\diff{\mathop{\backslash}}

\newcommand{\lexle}{{\mathchoice%
{\mathrel{\preceq\mkern-8.5mu\raise0.2ex\hbox{$\wr$}}}%
{\mathrel{\preceq\mkern-8.5mu\raise0.2ex\hbox{$\wr$}}}%
{\mathrel{\scriptstyle\preceq\mkern-3mu\raise0.1ex\hbox{$\scriptstyle\wr$}}}%
{\mathrel{\scriptscriptstyle\preceq\mkern-5mu\raise0.1ex\hbox{%
$\scriptscriptstyle\wr$}}}}}

\def\defeq{\mathrel{\buildrel \rm def \over =}}

\newcommand{\th}[1]{\mbox{$#1^{\rm th}\;$}}

\newenvironment{arblk}%
{\begin{array}[t]{l}}%
{\end{array}}

\newcommand{\LKP}{L^2_{\bms{K},\bms{P}}}
\newcommand{\bcomment}[1]{\mbox{\bf \hspace*{2em}#1}}

\begin{document}

\bibliographystyle{alpha}

\title{On Descriptive Complexity, Language Complexity, and GB\thanks{To Appear:
Specifying Syntactic Structures (papers from the Logic, Structures, and Syntax
workshop, Amsterdam, Sept. 1994)}}
\author{James Rogers\\
Institute for Research in Cognitive Science\\
University of Pennsylvania\\
Suite 400C, 3401 Walnut St\\
Philadelphia, PA 19104-6228, USA\\
{\tt jrogers@linc.cis.upenn.edu}}

\maketitle

\begin{abstract}
We introduce  $\LKP$, a monadic second-order language for reasoning
about trees which characterizes the strongly Context-Free Languages in the
sense that a set of finite trees is definable in $\LKP$ iff it is (modulo a
projection) a Local Set---the set of derivation trees generated by a CFG.
This provides a flexible approach to establishing language-theoretic complexity
results for formalisms that are based on systems of well-formedness constraints
on trees.  We demonstrate this technique by sketching two such results for
Government and Binding Theory.  First, we show that {\em free-indexation\/},
the mechanism assumed to mediate a variety of 
agreement and binding relationships in GB, is not definable in $\LKP$ and
therefore not enforcible by CFGs.  Second, we show how, in spite of this
limitation, a reasonably complete GB account of English can be defined in
$\LKP$.  
Consequently, the language licensed by that account is strongly context-free.
We illustrate some of the issues involved in establishing this result by
looking at the definition, in $\LKP$, of chains.  The limitations of this
definition provide some insight into the types of natural linguistic principles
that correspond to higher levels of language complexity.  We close with some
speculation on the possible significance of these results for generative
linguistics.
\end{abstract}

\section{Introduction}\label{sec.intro}
One of the more significant developments in generative linguistics over
the last decade has been the development of {\em
constraint-based} formalisms---grammar formalisms that define languages
not in terms of the derivations of the strings in the language, but
rather in terms 
of well-formedness conditions on the structures analyzing their syntax.
Because traditional notions 
of language complexity are generally defined in terms of rewriting mechanisms,
complexity of the languages licensed by these formalisms can be
difficult to determine.

A particular example, one that will be a focus of this paper, is
Government and Binding Theory.  While this is often modeled as a
specific range of Transformational Grammars, the connection between
the underlying grammar mechanism and the language a given GB theory
licenses is quite weak.  In an extreme view, one can take the underlying
mechanism simply to generate the set of all finite trees (labeled with
some alphabet of symbols)\footnote{Or, following
a strictly derivational approach, the set of all structures consisting of a
triple of finite trees along with a representation of PF.} 
while the grammatical theory is actually 
embodied in a set of principles that filter out the ill-formed analyses.
As a result, it has been difficult to establish language
complexity results for GB theories, even at the level of the
recursive~\cite{lapoin77,berwic84}
or context-sensitive~\cite{BerWei84} languages.

That language complexity results for GB should be difficult to come by
is hardly surprising.  The development of GB coincided with the
abandonment, by GB theorists, of the presumption that the traditional
language complexity classes would provide any useful characterization of
the human languages.  This followed, at least in part, from the recognition of
the fact that
the structural properties that characterize natural languages as a class may
well not be those that can be distinguished by existing language complexity
classes. 
There was a realization that the theory needed to be driven
by the regularities identifiable in natural languages, rather than
those suggested by abstract mechanisms.  Berwick characterized this
approach as aiming to ``discover the properties of natural languages
first, and then characterize them formally.''~\cite[pg. 100]{berwic84}

But formal language theory still has much to offer to generative
linguistics.  Language complexity provides one of the most useful
measures with which to compare languages and language formalisms. 
We have an array of results establishing the
boundaries of these classes, and, while many of the results do not seem
immediately germane to natural languages,
even seemingly artificial diagnostics
(like the copy language $\set{ww\mid w\in (ab)^*}$) can provide the basis
for useful classification results (such as Shieber's argument for the
non-context-freeness of Swiss-German~\cite{shiebe85}).
More importantly, characterization results for language
complexity classes tend to be in terms of the {\em structure}
of languages, and the structure of natural language, while hazy, is
something that can be studied more or less directly.  Thus there is a
realistic expectation of finding empirical evidence falsifying a given
hypothesis.  (Although such evidence may well be difficult to find, as
witnessed by the history of less successful attempts to establish results such
as Shieber's~\cite{PulGaz82,pullum84}.)
Further,
language complexity classes characterize, along one dimension, the
{\em types}
of resources necessary to parse or recognize a language.  Results of
this type for the class of human languages, then, make specific
predictions about the nature of the human language faculty, predictions
that, at least in principle, can both inform and be informed by progress
in uncovering the physical nature of that faculty.

In this paper we discuss a flexible and quite powerful approach to
establishing language complexity results for formalisms based on systems
of constraints on trees.  In Section~\ref{sec.LKP} we introduce a logical
language, $\LKP$, capable
of encoding such constraints lucidly.  The key merit of such an encoding
is the fact that sets of trees are definable in $\LKP$ if and only if
they are strongly context-free. Thus definability in $\LKP$
characterizes the strongly context-free languages.  This is our primary
result, and we develop it in Section~\ref{sec.character}.

We have used this technique to establish both inclusion and exclusion
results for a variety of linguistic principles within the GB
framework~\cite{rogers94}. 
In the remainder of the paper we demonstrate some of these.  In
Section~\ref{sec.nondef} we sketch a proof of the non-definability of
free-indexation, a mechanism that is nearly ubiquitous in GB theories. 
The consequence of this result is that languages that are licensed by
theories that necessarily employ free-indexation are outside of the
class of CFLs.  Despite the unavailability of free-indexation, we are
able to capture a mostly standard GB account of English within $\LKP$. 
Thus we are able to show that the language licensed by this particular
GB theory is strongly context-free. In Section~\ref{sec.chains} we illustrate
some of the issues 
involved in establishing this result, particularly in light of the
non-definability of free-indexation.  We
close, finally, with some
speculation on the possible significance of these results for generative
linguistics.

\section{$L^2_{{K},{P}}$}\label{sec.LKP}

The idea of employing mathematical logic to provide a precise formalization of
GB theories is a natural one.  This has been done, for instance, by
Johnson~\cite{johnso89} 
and Stabler~\cite{stable92} using first-order logic (or the Horn-clause
fragment of first-order logic) and by Kracht~\cite{kracht95} using a fragment
of dynamic logic.  What distinguishes the formalization we discuss
is the fact that it is
carried out in a language which can only define strongly context-free sets.
The fact that the formalization is possible, then, establishes a relatively 
strong language complexity result for the theory we capture.

We have, then, two conflicting criteria for our language.  It must be
expressive enough to capture the relationships that define the trees licensed
by the theory, but it must be restricted sufficiently to be no more
expressive than Context-Free Grammars.
In keeping with the first of these our language is intended to support, as
transparently as possible, 
the kinds of reasoning about trees typical of linguistic applications.
It includes binary predicates for  the usual structural relationships between
the nodes in the trees---parent (immediate domination), 
domination (reflexive), proper domination (irreflexive), left-of (linear
precedence) and equality.  In addition, it includes an arbitrary array
of monadic predicate constants---constants naming specific subsets of
the nodes in the tree.  These can be thought of as atomic labels.
The formula $\f{NP}(x)$, for instance, is true at
every node labeled \xp{N}.  It includes, also, a similar array of
individual constants---constants naming specific individuals in the
tree---although these prove to be of limited usefulness.  There are two
sorts of variables as well---those that range over nodes in the tree and those
that range over arbitrary subsets of those nodes (thus this is is monadic
second-order language).  Crucially, though, this is all the language includes.
By restricting ourselves to this language we restrict ourselves to working with
properties that can be expressed in terms of these basic predicates.

To be precise, the actual language we use in a given situation depends on the
sets of constants in use in that context.  We are concerned then with a
family of languages, parameterized by the sets of individual and set
constants they employ.

\begin{definition}\label{def.ax.lang}
For $\bm{K}$ a set of individual constant symbols,
and $\bm{P}$ a set of propositional constant symbols, both countable,
let $\LKP$ be the language built up from $\bm{K}$, $\bm{P}$,
a fixed countably infinite
set of {\em ranked} variables $\bm{X}=\bm{X}^0\cup\bm{X}^1$,
and the symbols:\\
\begin{tabular}{rcl}
$\parent,\dom,\prdom,\lft$  & --- &
    two place predicates, {\em parent}, {\em domination}, {\em proper
domination}\\
& & and {\em left-of} respectively,\\
$\eq$ & --- & equality predicate,\\
\multicolumn{3}{l}{$\wedge,\vee,\neg,\ldots,\forall,\exists,(,),[,]$ ---} \\
& & usual logical connectives, quantifiers, and grouping symbols.
\end{tabular}
\end{definition}

We use infix notation for the fixed predicate constants
$\parent$, $\dom$, $\prdom$, $\lft$, and $\eq$.  We use lower-case for
individual variables and constants, and upper-case for set variables and
predicate constants.  Further, we will say $X(x)$ to assert that the individual
assigned to the variable $x$ is included in the set assigned to the
variable $X$.  So, for instance,
\[
(\forall y)[x\dom y\limp X(y)]
\]
asserts that the set assigned to $X$ includes every node dominated by the node
assigned to $x$.

Truth, for these languages, is defined relative to a specific class of models.
The basic models are just ordinary structures interpreting the individual
and predicate constants.
{\fussy
\begin{definition}
A {\bf model} for the language $L_{\bms{K},\bms{P}}$
is a tuple $\tup{\sU,\sI,\sP,\sD, \sL, \sR_p}_{p\in \bms{P}}$, where:\newline
\begin{tabular}[t]{l}
$\sU$ is a non-empty universe,\\
$\sI$ is a function from $\bm{K}$ to $\sU$,\\
$\sP$, $\sD$, and $\sL$ are binary relations over $\sU$
    (interpreting $\parent$, $\dom$, and $\lft$ respectively),\\
$\sR_p$ is a subset of $\sU$ interpreting $p$.
\end{tabular}
\end{definition}}
If the domain of $\sI$ is empty (i.e., the model is for a language
$L_{\emptyset,\bms{P}}$) we will generally omit it.  Models for
$L_{\emptyset,\emptyset}$, then, are tuples $\tup{\sU,\sP,\sD,\sL}$.

The intended class of these models are, in essence, labeled {\em tree domains}.
A tree domain is the set of node addresses generated by giving
the address $\epsilon$ to the root and giving the children of the node
at address $w$ addresses (in order, left to right)
$w\cdot 0, w\cdot 1,\ldots$, where the centered dot denotes
concatenation.\footnote{We will usually dispense with the dot and denote
concatenation by juxtaposition.}
Tree domains, then, are particular subsets of $\Nat^*$.  ($\Nat$ is the
set of natural numbers.)

\begin{definition}
A {\bf tree domain} is a non-empty
set $T\subseteq\Nat^*$, satisfying, for all $u,v\in
\Nat^*$ and $i,j \in \Nat$, the conditions:
\[
\bm{TD1}\qquad uv\in T\IMP u\in T,\qquad\qquad
\bm{TD2}\qquad ui\in T,\; j<i \IMP uj\in T.
\]
\end{definition}

Every tree domain has a natural interpretation as a model for
$L_{\emptyset,\emptyset}$ (which interprets only the fixed predicate symbols.)
\begin{definition}\label{def.ax.natural}
The {\bf natural interpretation} of a tree domain $T$ is a model
$T^\natural=\tup{T,\sP^T,\sD^T,\sL^T}$,
where:
\[\begin{array}{rcl}
\sP^T &=& \set{\pair{u}{ui}\in T\times T\mid u\in\Nat^*, i\in\Nat},\\
\sD^T &=& \set{\pair{u}{uv}\in T\times T\mid u,v\in\Nat^*},\\
\sL^T &=&
    \set{\pair{uiv}{ujw}\in T\times T\mid u,v,w\in\Nat^*, i<j\in\Nat}.
\end{array}
\]
\end{definition}

The structures of interest to us are just those models that are the natural
interpretation of a tree domain, augmented with interpretations of
additional individual and predicate constants.\footnote{A partial
axiomatization of this class of models is given in~\cite{rogers94}.}

In general, satisfaction is relative to an assignment mapping each
individual variable into a member of $\sU$ and each predicate variable
into a subset of $\sU$.  We use
\[
M\models\phi\,[s]
\]
to denote that a model $M$ satisfies a formula $\phi$ with an
assignment $s$.  The notation
\[
M\models\phi
\]
asserts that $M$ models $\phi$ with any assignment.  When $\phi$ is a
sentence (has no unquantified variables) we will usually use this form.

Proper domination is a defined predicate:
\[
M\models x\prdom y\,[s] \IFF M\models x\dom y,x\not\eq  y\,[s].
\]

\subsection{Definability in $L^2_{K,P}$}\label{sec.define}

We are interested in the subsets of the class of intended
models which are definable
in $\LKP$ using any sets $\bm{K}$ and $\bm{P}$.  
If  $\Phi$ is a set of
sentences in a language $\LKP$, we will use the notation $\Mod(\Phi)$ to denote
the set of {\em trees}, i.e., 
intended models, that satisfy all of the sentences in $\Phi$.  We are
interested, then, in the sets of trees that are $\Mod(\Phi)$ for some such
$\Phi$.   In developing our definitions we can use individual and monadic
predicates freely (since $\bm{K}$ and $\bm{P}$ can always be taken to
be the sets that actually occur in our definitions) and we can quantify over
individuals and sets of individuals.  We will also use
non-monadic predicates and even higher-order predicates,
e.g., properties of subsets, but only those that can be {\em
explicitly} defined, that is, those which can be eliminated by a simple
syntactic replacement of the predicate by its definition.

This use of explicitly defined predicates is crucial to the transparency of
definitions in $\LKP$.  We might, for instance, define a simplified version of
government in three steps:
\begin{eqnarray*}
\f{Branches}(x) &\liff& 
(\exists y,z)[x\parent y\land x\parent z\land y\not\eq z]\\
\f{C-Command}(x,y) & \equiv & \neg x\dom y \land \neg y\dom x\land
(\forall z)[(z\prdom x\land \f{Branches}(z))\limp z\prdom y]\\
\f{Governs}(x,y) & \equiv&
\f{C-Commands}(x,y)\land\\
&&\neg(\exists z)[\f{Barrier}(z)\land z\prdom y\land
\neg z\prdom x],
\end{eqnarray*}
in words, $x$ governs $y$ iff it c-commands $y$ and no barrier
intervenes between them.  It 
c-commands $y$ iff neither $x$ nor $y$ dominates the other and every
branching node that properly dominates $x$ also properly dominates $y$.
$\f{Branches}(x)$ is just a monadic predicate; it is within the language of
$\LKP$ (for suitable $P$) and its definition is simply a biconditional $\LKP$
formula. 
In contrast, C-Command and Governs are non-monadic and do
not occur in $\LKP$.  Their definitions, however, are ultimately in
terms of monadic predicates and the fixed predicates (parent,
etc.) only. One can replace each of their occurrences in a formula
with the right hand side of their definitions and eventually derive a
formula that {\em is} in $\LKP$.  We will reserve the use of $\equiv$ (in
contrast to $\liff$) for explicit definitions of non-monadic predicates.

Definitions can also use predicates expressing properties of sets and relations
between sets, as long as those properties can be explicitly defined.  The 
subset relation, for instance can be defined:
\[
\f{Subset}(X,Y) \equiv (\forall x)[X(x)\limp Y(x)].
\]
We can also
capture the stronger notion of one set being partitioned by a collection
of others:
\[
\f{Partition}(\vec{X},Y)\equiv
(\forall x)\left[\left(Y(x)\liff\bigvee_{X\in\vec{X}}X(x)\right)\,\land\,
\bigwedge_{X\in\vec{X}}\left[X(x)\limp\bigwedge_{Z\in\vec{X}\diff\set{X}}
\neg Z(x)\right]\right].
\]
Here $\vec{X}$ is a some sequence of set variables and 
$\bigvee_{X\in\vec{X}} X(x)$ is shorthand for the disjunction $X_0(x)\lor
X_1(x)\cdots$ for all $X_i$ in $\vec{X}$, etc.  There is a distinct
instance of $\f{Partiton}$ for each sequence $\vec{X}$, although we can ignore
distinctions between sequences of the same length.
Finally, we note that finiteness is a definable property of subsets in
our intended models.  This follows from the fact that these models are
linearly ordered by the  {\em lexicographic order} relation: 
\[
x\lexle y \equiv x\dom y\lor x\lft y.
\]
and that every non-empty subset of such a model has a least element with
respect to that order.  
A set of nodes, then, is finite iff
each of its non-empty subsets has an upper-bound with respect to lexicographic
order as well.
\[
\f{Finite}(X)\equiv
(\forall Y)[(\f{Subset}(Y,X)\land(\exists x)[Y(x)])\limp
(\exists x)[Y(x)\land(\forall y)[Y(y)\limp y\lexle x]]].
\]
These three second-order relations will play a role in the next section.

\section{Characterizing the Local Sets}\label{sec.character}
We can now give an example of a class of sets of trees that is definable in
$\LKP$---the local sets (i.e., the sets of derivation trees generated by
Context-Free Grammars).  The idea behind the definition is simple.  Given an
arbitrary Context-Free Grammar, we can treat its terminal and
non-terminal symbols as monadic predicate constants.  The productions of the
grammar, then, relate the label of a node to the number and labels of its
children.  If the set of productions for a non-terminal $A$, for instance, is
\[
\f{A}\longrightarrow \f{Bc} \mid \f{AB} \mid \f{d}
\]
we can translate this as
\[
\begin{array}{ll}
(\forall x)[\f{A}(x)\limp(&
(\exists y_1,y_2)[\f{Children}(x,y_1,y_2)\land\f{B}(y_1)\land\f{c}(y_2)]\lor\\
&(\exists y_1,y_2)[\f{Children}(x,y_1,y_2)\land\f{A}(y_1)\land\f{B}(y_2)]\lor\\
&(\exists y_1)[\f{Children}(x,y_1)\land\f{d}(y_1)]\quad),
\end{array}
\]
where
\[
\begin{array}{rcl}
\f{Children}(x,y_1,\ldots,y_n)&\equiv&
\bigwedge_{i\leq n}[x\parent y_i]\land
\bigwedge_{i<j\leq n}[y_i\lft y_j]\land\\
&&(\forall z)[x\parent z\limp\bigvee_{i\leq n}[z\eq y_i]].
\end{array}
\]
We can collect such translations of all the productions of the grammar together
with sentences requiring nodes labeled with terminal symbols to have no
children, requiring the root to be labeled with the start symbol, requiring
the sets of nodes labeled with the terminal and non-terminal symbols to
partition the set of all nodes in the tree, and requiring that set of nodes to
be finite.  It is easy to show that the models of this set of sentences
are all and only the derivation trees of the grammar.\footnote{A more complete
proof is given in~\cite{rogers94}.}  In this way we get the first half
of our characterization of the local sets.

\begin{theorem}\label{thm.definable}
The set of derivation trees generated by an arbitrary Context-Free Grammar is
definable in $\LKP$.
\end{theorem}

It is, perhaps, not surprising that we can define the local sets with $\LKP$.
This is superficially quite a powerful language, allowing, as it does, a
certain amount of second-order quantification.  It is maybe more remarkable
that, modulo a projection, the {\em only} sets of finite trees (with
bounded branching)
that are definable in $\LKP$ are the local sets.  

\begin{theorem}\label{thm.character}
Every set of finite trees with bounded branching 
that is definable in $\LKP$ is
the projection of a set of trees generated by a finite set of Context-Free
(string) Grammars.
\end{theorem}

The proof hinges on the fact that one can translate formulae in $\LKP$ into the
language of S$n$S---the monadic second-order theory of multiple successor
functions.  This is the monadic second-order theory of the structure
\[
\sN_n\defeq\tup{T_n,\dom,\lexle,\f{r}_i}_{i<n},
\]
a generalization of the
natural numbers with successor and less-than.  The universe, $T_n$, is the
complete $n$-branching tree domain.  The relation $\dom$ is domination,
$\lexle$ is lexicographic order, and the functions $\f{r}_i$ are the
successor functions, each taking nodes into their $\th{i}$ child ($w\mapsto
wi$).  Rabin~\cite{rabin69} showed that S$n$S is decidable for any $n\leq
\omega$.  One way of understanding his proof is via the observation that 
satisfying assignments for
a formula $\phi(\vec{X})$, with free variables\footnote{We
will assume, for simplicity, that only set variables occur free.  Since
individual variables can be re-interpreted as variables ranging over
singleton sets, this is without loss of generality.} among $\vec{X}$ 
can be understood as trees
labeled with (subsets of) 
the variables in $\vec{X}$.  A node is in the set assigned to
$X_i$ in $\vec{X}$ iff it is labeled with $X_i$.  Rabin showed that, for any
$\phi(\vec{X})$ in the language of S$n$S, the set of trees encoding the
satisfying assignments for $\phi(\vec{X})$ in $\sN_n$ is accepted by a
particular type of finite-state automaton on infinite trees.  We say that the
set is {\em Rabin recognizable}.  He goes on to show that emptiness of these
sets is decidable.  It follows that satisfiability of these
formulae, and hence the theory S$n$S, is decidable.

For us, the key point is the fact that the sets encoding satisfying assignments
are Rabin recognizable.  It is not difficult to exhibit a syntactic
transformation which, given any $\psi(\vec{X})$ in $\LKP$, produces a
formula $\phi(X_U,\vec{X}_P,\vec{X})$ in the language of S$n$S,
where $X_U$ is a new variable and
$\vec{X}_P$ is a sequence of new variables (one for each of the finitely many
predicates in \bm{P} that occur in $\psi$) such that,
\[
\sN_n\models\phi[A_U,\vec{A}_P,\vec{A}]
\]
iff
\[
\tup{A_U,\sP^{A_U},\sD^{A_U},\sL^{A_U},\vec{A}_P}\models\psi[\vec{A}],
\]
that is, the set $A_U$ and 
the sequences of sets $\vec{A}_P$ and $\vec{A}$ form
a satisfying assignment for $\phi$ in $\sN_n$ iff the structure consisting of
the universe $A_U$ along with the natural interpretation of $\parent$, 
$\dom$,
and $\lft$ on $A_U$, and the sets $\vec{A}_P$, 
satisfies $\psi$ with the assignment taking $\vec{X}$
into $\vec{A}$.  It follows that a set of trees is definable in $\LKP$ iff
they are Rabin recognizable.

If we restrict our attention to sets of finite trees, we can take Rabin's
automata
to be ordinary finite-state automata over finite trees~\cite{GecSte84}, that
is, 
the sets of finite trees that are definable in $\LKP$ are simply {\em
recognizable}.  One can think of these automata as traversing the tree, top
down, assigning states to the children of a node on the basis of a transition
function that depends on the state of the node, its label, and the position of
the child among its siblings.  A tree is accepted if it can be labeled by the
automaton in such a way that the root is labeled with a start state and the
set of states labeling the leaves is one of a set of accepting sets of
states. Every set of trees that is accepted in this way is the projection of a
local set.   To see this,\footnote{This
proof is evidently originally due to Thatcher~\cite{thatch67}.  In addition,
Theorem~\ref{thm.character} is implicit in the proof of a related theorem due to
Doner~\cite{doner70}.} suppose that
$\tau$ is a tree accepted by a tree automaton.  Then there is some assignment
of states to the nodes in $\tau$ that witnesses this fact.  
\begin{figure}
\begin{center}
\begin{picture}(0,0)%
\includegraphics{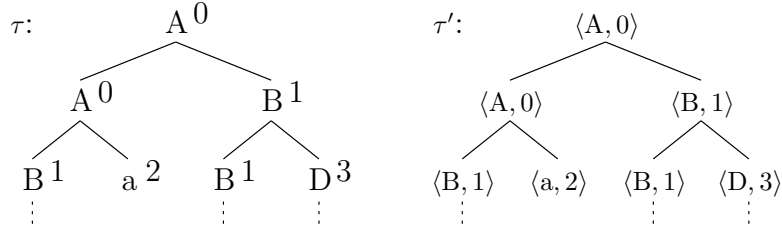}%
\end{picture}%
\setlength{\unitlength}{0.012500in}%
\begin{picture}(379,96)(-14,740)
\put( 80,823){\makebox(0,0)[b]{\raisebox{0pt}[0pt][0pt]{\twlrm A}}}
\put( 40,790){\makebox(0,0)[b]{\raisebox{0pt}[0pt][0pt]{\twlrm A}}}
\put(120,790){\makebox(0,0)[b]{\raisebox{0pt}[0pt][0pt]{\twlrm B}}}
\put( 20,757){\makebox(0,0)[b]{\raisebox{0pt}[0pt][0pt]{\twlrm B}}}
\put(100,757){\makebox(0,0)[b]{\raisebox{0pt}[0pt][0pt]{\twlrm B}}}
\put(140,757){\makebox(0,0)[b]{\raisebox{0pt}[0pt][0pt]{\twlrm D}}}
\put( 60,757){\makebox(0,0)[b]{\raisebox{0pt}[0pt][0pt]{\twlrm a}}}
\put( 90,827){\makebox(0,0)[b]{\raisebox{0pt}[0pt][0pt]{\twlrm 0}}}
\put( 50,794){\makebox(0,0)[b]{\raisebox{0pt}[0pt][0pt]{\twlrm 0}}}
\put(130,794){\makebox(0,0)[b]{\raisebox{0pt}[0pt][0pt]{\twlrm 1}}}
\put(110,761){\makebox(0,0)[b]{\raisebox{0pt}[0pt][0pt]{\twlrm 1}}}
\put( 30,761){\makebox(0,0)[b]{\raisebox{0pt}[0pt][0pt]{\twlrm 1}}}
\put( 70,761){\makebox(0,0)[b]{\raisebox{0pt}[0pt][0pt]{\twlrm 2}}}
\put(150,761){\makebox(0,0)[b]{\raisebox{0pt}[0pt][0pt]{\twlrm 3}}}
\put(260,823){\makebox(0,0)[b]{\raisebox{0pt}[0pt][0pt]{\twlrm $\pair{\f{A}}{0}$}}}
\put(220,790){\makebox(0,0)[b]{\raisebox{0pt}[0pt][0pt]{\twlrm $\pair{\f{A}}{0}$}}}
\put(300,790){\makebox(0,0)[b]{\raisebox{0pt}[0pt][0pt]{\twlrm $\pair{\f{B}}{1}$}}}
\put(200,757){\makebox(0,0)[b]{\raisebox{0pt}[0pt][0pt]{\twlrm $\pair{\f{B}}{1}$}}}
\put(280,757){\makebox(0,0)[b]{\raisebox{0pt}[0pt][0pt]{\twlrm $\pair{\f{B}}{1}$}}}
\put(320,757){\makebox(0,0)[b]{\raisebox{0pt}[0pt][0pt]{\twlrm $\pair{\f{D}}{3}$}}}
\put(240,757){\makebox(0,0)[b]{\raisebox{0pt}[0pt][0pt]{\twlrm $\pair{\f{a}}{2}$}}}
\put( 20,823){\makebox(0,0)[rb]{\raisebox{0pt}[0pt][0pt]{\twlrm $\tau$:}}}
\put(200,823){\makebox(0,0)[rb]{\raisebox{0pt}[0pt][0pt]{\twlrm $\tau'$:}}}
\end{picture}
\end{center}
\caption{Proof of Theorem~\protect{\ref{thm.character}}\label{fig.prop17}}
\end{figure}
Suppose, for
instance, $\tau$ is the tree of Figure~\ref{fig.prop17}, labeled as shown.
Consider the tree $\tau'$ in which each node is labeled with a pair consisting
of the label from 
$\tau$ and the state assigned to that node.  It is easy to show that, given a
recognizable set of trees, one can construct a CFG to generate the
corresponding set of trees labeled with pairs as in $\tau'$.  In the example,
for instance, this would include, among others, the productions
\[
\begin{array}{rcl}
\pair{\f{A}}{0}&\longrightarrow& \pair{\f{A}}{0}\pair{\f{B}}{1}
\mid\pair{\f{B}}{1}\pair{\f{a}}{2}\mid\cdots\\
\pair{\f{B}}{1}&\longrightarrow& \pair{\f{B}}{1}\pair{\f{D}}{3}\mid\cdots\\
&\vdots&
\end{array}
\]
The original set of trees is then the first projection of the set generated by
the CFG.

Together, these two theorems give us our primary result.
\begin{corollary}
A set of finite trees with bounded branching is local (modulo projection) iff
it is definable in $\LKP$.
\end{corollary}

\section{Non-Definability of Free Indexation}\label{sec.nondef}
This characterization provides a powerful tool for establishing strong
context-freeness of  classes of languages that are
defined by constraints on the structure of the trees analyzing the strings in
the language.  If one can show that the
constraints defining such a set, or perhaps that any constraints in the class
employed by a given formalism, can be defined within
$\LKP$ then the corresponding language or class of languages is
strongly context-free.  Much of the value of standard language complexity
classes, on the other hand, comes from results that allow one to show that a
given language or class of languages is not included in a particular complexity
class.  Such
results are available here as well, in the form of non-definability results for
$\LKP$.  One relatively easy way of establishing such results is by employing
the contrapositive of Theorem~\ref{thm.character}.  If one can
show that a given predicate, when added to $\LKP$ allows definition of known
non-CF languages, then clearly that predicate properly extends the power of the
language and cannot be definable.  In this way, one can show that the predicate
$\f{YieldsEq}_P(x,y)$ which holds between two nodes iff the yields of the
subtrees rooted at those nodes are labeled identically wrt $P$ is not definable
in $\LKP$, for if it were one could define the copy language $\set{ww\mid w\in
(ab)^*}$.

In this section we will explore an approach that is more difficult but is
one of the most  general---reduction from the monadic second-order theory of
the grid---and will use it to demonstrate non-definability of
free-indexation---a mechanism which shows up in a number of modules of GB.

The grid is the structure
$G=\tup{\Nat^2,\f{O},\f{r}_0,\f{r}_1}$ where
\begin{eqnarray*}
\f{O} & = & \pair{0}{0}\\
\f{r}_0(\pair{x}{y}) &=& \pair{x+1}{y}\\
\f{r}_1(\pair{x}{y}) &=& \pair{x}{y+1}.
\end{eqnarray*}
This is the structure of the (discrete) first quadrant.  Note the similarity to
$\sN_2$, the structure of two successor functions.  The key distinction is the
fact that $G$ satisfies the property 
\[
(\forall x)[\f{r}_0(\f{r}_1(x))=\f{r}_1(\f{r}_0(x))],
\]
that is, the horizontal successor of the vertical successor of a point is the
same as the vertical successor of its horizontal successor.  Let $\Th_2(G)$ be
the monadic second-order theory of $G$.  Lewis~\cite{lewis79} showed that
this theory is undecidable by showing how one could define the set of
terminating computations of an arbitrary Turing machine within it.

Now, the monadic second-order theory of any of our intended structures is
decidable (by reduction to S$n$S), as is the monadic second-order theory of any
of our intended structures augmented with any predicate that is definable in
$\LKP$ (since we can reduce this to the theory of the original structure via
that definition).  Our approach to showing that a predicate is not
definable in $\LKP$ is to show that the theory of one of our structures
augmented with that predicate is not decidable.  In particular, we will show
that the theory of such a structure includes an undecidable fragment of the
monadic second-order theory of the grid.

Our focus, in this section, is the mechanism known as {\em free-indexation}.
In the Government and Binding Theory framework this is the mechanism that
is generally assumed
to mediate issues like agreement, co-reference of nominals, and identification
of moved elements with their traces.  In its most general form this operates by
assigning indices to the nodes of the tree randomly and then filtering out
those assignments that do not meet various constraints on agreement,
co-reference, etc.  In essence, the indexation is an equivalence relation, one
that distinguishes  
unboundedly many equivalence classes among the nodes of the tree.  That is,
each value of the index identifies an equivalence class and there is no a
priori bound on its maximum value.  Free-indexation views constraints on the
indexation as a filter that admits only those equivalence relations that meet
specific conditions on the relationships between the individuals in these
classes.

To see that we cannot define such equivalence relations in $\LKP$, consider the
class of structures
\[
\sT_\f{CI}=\tup{T_2,\sP_2,\sD_2,\sL_2,\f{CI}},
\]
where $T_2$ is the complete binary-branching tree domain, $\sP_2$, $\sD_2$, and
$\sL_2$ are the natural interpretations of parent, domination, and left-of on
that domain, and $\f{CI}$ is any arbitrary equivalence relation.   Let S2S+CI
be the monadic second-order theory of this class of structures.
Our claim is that
this is an undecidable theory.\footnote{Since the property of being an
equivalence relation---being reflexive, symmetric, and transitive---is definable
in $\LKP$, our result is one way of showing that $\sN_2$ augmented with a
single arbitrary binary relation has a non-decidable monadic second-order
theory.}

\begin{theorem}
S2S+CI is not decidable.
\end{theorem}

Lewis's proof of the non-decidability of $\Th_2(G)$ is based on a construction
that takes any given Turning Machine $M$ into a formula $\phi_M(\vec{P})$ 
such that $G\models(\exists \vec{P})[\phi_M(\vec{P})]$ iff $M$ halts (when
started, say, on the empty tape).  
The idea behind our proof of the non-decidability of S2S+CI is that there is a
natural correspondence between points in $T_2$ and those in $\Nat^2$ that is
induced by interpreting node addresses in $T_2$ as paths (non-decreasing in
both $x$ and $y$) from the origin in $\Nat^2$.  Of course, in general, there
will be many points in $T_2$ that correspond to the same point in $\Nat^2$, but
we can restrict the interpretation of $\f{CI}$ in such a way that all points in
$T_2$ that correspond to the same point in $\Nat^2$ will be co-indexed.  We
then restrict the interpretation of the variables in $\vec{P}$ in such a
way that it does not break the classes of $\f{CI}$.  In more typically
linguistic terms, we require 
co-indexed nodes to agree on the features in $\vec{P}$.

The formula
$\phi_M(\vec{P})$ of Lewis' proof
involves only the constant $\f{O}$, the successor functions
$\f{r}_0$ and $\f{r}_1$, some set of (bound) individual variables, the (free)
monadic predicate variables in $\vec{P}$, and the logical connectives.

Let
\begin{eqnarray*}
\f{O}(x) &\liff& (\forall y)[y\dom x\limp y\eq x]\\
\f{r}_0(x,y) &\equiv& x\parent y\land(\forall z)[x\parent z\limp z\not\lft y]\\
\f{r}_1(x,y) &\equiv& x\parent y\land(\forall z)[x\parent z\limp y\not\lft z].
\end{eqnarray*}
Then $\f{O}(x)$ is true only at the root,
$\f{r}_0(x,y)$ is true iff $y$ is the leftmost child of $x$ and
$\f{r}_1(x,y)$ is true iff $y$ is the rightmost child of $x$.  These
translations are sufficient for us to translate $\phi_M(\vec{P})$ into a
formula $\psi_M(\vec{P})$ that, when combined with an axiom
$\Phi_G(\vec{P})$
constraining the interpretation of $\f{CI}$  and $\vec{P}$ as sketched
above, will be satisfiable by a model in the class $\sT_\f{CI}$ iff
$\phi_M(\vec{P})$ is satisfied by $G$.  That is:
\[
\mbox{ There exists } T\in\sT_\f{CI}\mbox{ such that }
T\models(\exists\vec{P})[\psi_M(\vec{P})\land\Phi_G(\vec{P})]
\]
iff
\[
G\models(\exists\vec{P})[\phi_M(\vec{P})].
\]
This in turn implies that
\[
(\exists\vec{P})[\phi_M(\vec{P})]\in\Th_2(G)\quad\tiff\quad
\neg(\exists\vec{P})[\psi_M(\vec{P})\land\Phi_G(\vec{P})]\not\in
{\rm S2S+CI}.
\]
Decidability of S2S+CI, then, would imply decidability of the halting problem.

It remains only to define $\Phi_G(\vec{P})$.  Let
\begin{subeqnarray}
\lefteqn{\Phi_G(\vec{P}) \equiv}\nonumber\\
&(\forall x,y)[&
  \f{CI}(x,y)\lif(\begin{arblk}
    x\eq y \quad \lor\\
\lefteqn{\bcomment{---$x$ and $y$ are equal or}}\\
    (\exists x_0,y_0)[\begin{arblk}
      \f{CI}(x_0,y_0)\land\\
      (\begin{arblk}
	(\f{r}_0(x_0,x)\land\f{r}_0(y_0,y))\lor\\
	(\f{r}_1(x_0,x)\land\f{r}_1(y_0,y))\phantom{\lor})\,]\quad \lor
      \end{arblk}
    \end{arblk}\\
\lefteqn{\bcomment{---$x$ and $y$ are both left-children or both}}\\
\lefteqn{\bcomment{\hphantom{---}right-children of co-indexed nodes or}} \\
    (\exists x_0,y_0,x_1,y_1)[\begin{arblk}
      \f{CI}(x_0,y_0)\land\\
      \f{r}_0(x_0,x_1)\land\f{r}_1(x_1,x)\land\\
      \f{r}_1(y_0,y_1)\land\f{r}_0(y_1,y)\phantom{\land}\;]
    \end{arblk}\\
\lefteqn{\bcomment{---$x$ is the right-child of the left-child}}\\
\lefteqn{\bcomment{\hphantom{---}and $y$ is the left-child of the
right-child}}\\ 
\lefteqn{\bcomment{\hphantom{---}of co-indexed nodes, or v.v.}}\\
  \;)\quad\land\\
  \end{arblk}\slabel{eq.PhiGa}\\
  &&\f{CI}(x,y)\limp \f{Agree}_{\vec{P}}(x,y)\;]\slabel{eq.PhiGb},
\end{subeqnarray}
where
\[
\f{Agree}_{\vec{P}}(x,y)\equiv \bigwedge_{P\in\vec{P}}(P(x)\liff P(y)).
\]

This requires that
every node is co-indexed with itself,
that the left children of co-indexed nodes are
co-indexed as are the right children of co-indexed nodes,
and that the left child of the right child and right child of the left child of
co-indexed nodes are co-indexed.  Finally all co-indexed nodes are
forced, by $\f{Agree}_{\vec{P}}$,
to agree on all predicates in $\vec{P}$.
That this is sufficient to carry the reduction of the halting
problem to membership in S2S+CI depends on the fact that
$\Phi_G(\vec{P})$ forces all points in $T_2$ equivalent in the sense that
they correspond to the same
point in $G$ as sketched above, to agree on the predicates in $\vec{P}$. 
Thus we (roughly) can take the quotient with respect to
this equivalence without affecting satisfiability of
$\psi_M(\vec{P})$.  The resulting structure is isomorphic to $G$ and
satisfies $(\exists \vec{P})[\psi_M(\vec{P})]$ iff $G$ satisfies 
$(\exists \vec{P})[\phi_M(\vec{P})]$. 
The proof is carried out in detail in~\cite{rogers94}.

The non-definability of free-indexation is a significant obstacle to
capturing GB accounts of language in $\LKP$.  
As it turns out,  other constraints employed in GB theories are not generally
difficult to define.  Our ability to capture these accounts,
then, depends directly on the degree to which they necessarily employ
free-indexation.  
The common practice, in GB, is to simply assume co-indexation
almost whenever there is a 
need to identify components of the tree in some way.  Unfortunately, we
cannot capture directly accounts that are defined in these terms.
Rather, we are compelled to restate them
without reference to indices.  On the other hand, it is not at all
clear that accounts that appeal to free-indexation actually require so
general a mechanism.  On the contrary, it seems that indices are frequently
only a conceptually simple way of encoding more complicated, but less
general relationships.
There has been a tendency, in the more recent GB literature, to avoid
free-indexation in favor of these more specific relationships.  Chomsky, 
for instance, comments:
\begin{quotation}
A theoretical apparatus that takes indices seriously as entities\ldots
is questionable on more general grounds. Indices are basically the
expression of a relationship, not entities in their own right.  They
should be replaceable without loss by a structural account of the
relation they annotate.~\cite[pg. 49, note 52]{chomsk93}
\end{quotation}

This quote comes in the context of a suggestion for a re-interpretation
of the standard account of Binding Theory in a manner that avoids use
of indices.    Rizzi, in~\cite{rizzi90}, motivated by an examination of
a wide variety of extraction phenomena,
offers a re-interpretation of the Empty Category Principle and the
theory of chains that restricts the role of indices to a relatively
small class of movements.
As we will see in the next section,
Rizzi's theory provides us with the
foundation we need to capture a largely complete GB account of English
in $\LKP$.  We thus establish that this account licenses a strongly
context-free language.  It seems noteworthy that GB theorists have been led,
by purely linguistic considerations, to precisely the kind of re-interpretation
of the theory we require in order to establish our language-theoretic
results. 

\section{Defining Chains}\label{sec.chains}
We turn now to an example that is particularly relevant to the issue of
capturing a Government and Binding Theory account of English in $\LKP$, and in
particular capturing it without use of 
indices.  This is our definition of {\em chains}---the core notion in
contemporary GB accounts of movement.  Our exposition is intended to be
accessible without prior familiarity with GB, although possibly
mysterious in some of its details.  It will necessarily be
somewhat meager both in the details of the definition and in the details of the
underlying theory.  A more complete treatment can be found 
in~\cite{rogers94}.

\begin{figure}
\begin{center}
\begin{picture}(0,0)%
\includegraphics{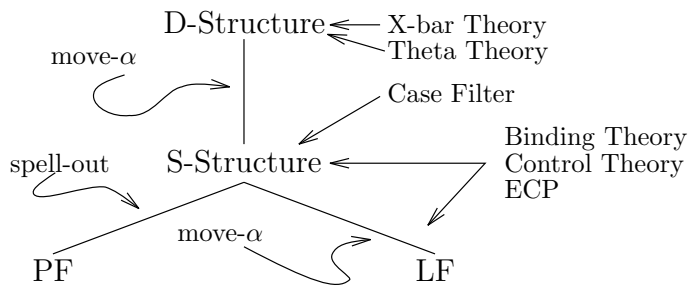}%
\end{picture}%
\setlength{\unitlength}{0.012500in}%
\begin{picture}(272,115)(62,696)
\put(269,755){\makebox(0,0)[lb]{\raisebox{0pt}[0pt][0pt]{\tenrm Binding Theory}}}
\put(269,744){\makebox(0,0)[lb]{\raisebox{0pt}[0pt][0pt]{\tenrm Control Theory}}}
\put(269,733){\makebox(0,0)[lb]{\raisebox{0pt}[0pt][0pt]{\tenrm ECP}}}
\put(160,802){\makebox(0,0)[b]{\raisebox{0pt}[0pt][0pt]{\twlrm D-Structure}}}
\put(160,744){\makebox(0,0)[b]{\raisebox{0pt}[0pt][0pt]{\twlrm S-Structure}}}
\put( 80,698){\makebox(0,0)[b]{\raisebox{0pt}[0pt][0pt]{\twlrm PF}}}
\put(240,698){\makebox(0,0)[b]{\raisebox{0pt}[0pt][0pt]{\twlrm LF}}}
\put(220,802){\makebox(0,0)[lb]{\raisebox{0pt}[0pt][0pt]{\tenrm X-bar Theory}}}
\put(220,791){\makebox(0,0)[lb]{\raisebox{0pt}[0pt][0pt]{\tenrm Theta Theory}}}
\put(220,773){\makebox(0,0)[lb]{\raisebox{0pt}[0pt][0pt]{\tenrm Case Filter}}}
\put( 79,789){\makebox(0,0)[lb]{\raisebox{0pt}[0pt][0pt]{\tenrm move-$\alpha$}}}
\put( 62,744){\makebox(0,0)[lb]{\raisebox{0pt}[0pt][0pt]{\tenrm spell-out}}}
\put(132,715){\makebox(0,0)[lb]{\raisebox{0pt}[0pt][0pt]{\tenrm move-$\alpha$}}}
\end{picture}
\end{center}
\caption{Levels of representation.\label{fig.levels}}
\end{figure}
\subsection{Identifying Antecedents of Traces}
Government and Binding Theory analyzes sentences with four distinct
syntactic representations which are related by the general transformation
{\em move-$\alpha$}.  These are {\em D-Structure}---corresponding
to the deep-structure of earlier transformational theories, {\em
S-Structure}---roughly corresponding to the surface-structure of those
theories, {\em Phonetic Form}---the actual phonetic structure of the sentence,
and {\em Logical Form}---a more or less direct representation of the sentence's
semantic content.  The principles embodying a GB theory of
language are collected into modules which apply at various levels of
this analysis. 
The principles
we capture include basic X-bar Theory, Theta Theory, the Case Filter, 
Binding Theory, Control Theory and various constraints on movement, in
particular the Empty Category Principle.
In this section we focus on the Empty Category Principle and the
definition of {\em chains}.

As we noted in the introduction, we prefer to regard GB theories as a
set of constraints on structures rather than a mechanism for
constructing them.  We take this a step further by assuming that those
constraints apply to a single tree which includes S-Structure and
D-Structure as submodels,\footnote{While we don't treat Logical Form,
there is no reason this cannot be incorporated into our structures in
much the same way.} rather than having some constraints apply to one
structure, others to the other, and others still to the relationship
between them.  In this view, D-Structure and move-$\alpha$ are best
understood as perspicuous means of stating constraints which are
obscured in a single-level representation (see, for instance,
Koster~\cite{koster87} and Brody~\cite{brody93}).\footnote{It is
interesting that Johnson, in~\cite{johnso89} initially defines all
four levels of structure, but then, through a series of standard
program transformations, optimizes away everything except PF and LF.}
One argument against such a view is that in some cases (such as
head-raising) chains formed by one movement can be disrupted by
subsequent movement.  Indeed, representational accounts, such as ours,
frequently appeal to a notion of {\em reconstruction}---effectively
derivation in reverse---to resolve such difficulties.  In fact, at
least if one can employ indices to identify the elements of chains,
there is no need for such a retreat.  Even limiting oneself to the
language of $L^2_{K,P}$, if one restricts attention to languages, like
English, in which head-movement is strictly limited, it is possible to
get a purely declarative (and reasonably clear) account of the issues
usually treated by reconstruction.  Details of such an account are
given in~\cite{rogers94}.

\begin{figure}
\begin{center}
\begin{picture}(0,0)%
\includegraphics{ecp1.pstex}%
\end{picture}%
\setlength{\unitlength}{0.012500in}%
\begin{picture}(441,217)(10,611)
\put( 66,817){\makebox(0,0)[b]{\raisebox{0pt}[0pt][0pt]{\elvrm \xp{C}}}}
\put(103,770){\makebox(0,0)[b]{\raisebox{0pt}[0pt][0pt]{\elvrm C}}}
\put( 66,770){\makebox(0,0)[b]{\raisebox{0pt}[0pt][0pt]{\elvrm I\sub{j}}}}
\put(121,770){\makebox(0,0)[b]{\raisebox{0pt}[0pt][0pt]{\elvrm \xp{N}}}}
\put(158,755){\makebox(0,0)[b]{\raisebox{0pt}[0pt][0pt]{\elvrm t\sub{j}}}}
\put(231,708){\makebox(0,0)[b]{\raisebox{0pt}[0pt][0pt]{\elvrm t\sub{i}}}}
\put(341,660){\makebox(0,0)[b]{\raisebox{0pt}[0pt][0pt]{\elvrm I}}}
\put(396,660){\makebox(0,0)[b]{\raisebox{0pt}[0pt][0pt]{\elvrm \xp{V}}}}
\put(396,645){\makebox(0,0)[b]{\raisebox{0pt}[0pt][0pt]{\elvrm \xb{V}}}}
\put(433,630){\makebox(0,0)[b]{\raisebox{0pt}[0pt][0pt]{\elvrm t\sub{i}}}}
\put( 48,786){\makebox(0,0)[b]{\raisebox{0pt}[0pt][0pt]{\ninrm Whom}}}
\put( 66,755){\makebox(0,0)[b]{\raisebox{0pt}[0pt][0pt]{\ninrm do}}}
\put(103,755){\makebox(0,0)[b]{\raisebox{0pt}[0pt][0pt]{\ninrm $\emptyset$}}}
\put(121,755){\makebox(0,0)[b]{\raisebox{0pt}[0pt][0pt]{\ninrm you}}}
\put(378,614){\makebox(0,0)[b]{\raisebox{0pt}[0pt][0pt]{\ninrm invite}}}
\put(108,802){\makebox(0,0)[b]{\raisebox{0pt}[0pt][0pt]{\elvrm \xb{C}}}}
\put( 79,786){\makebox(0,0)[b]{\raisebox{0pt}[0pt][0pt]{\elvrm C}}}
\put(181,770){\makebox(0,0)[b]{\raisebox{0pt}[0pt][0pt]{\elvrm \xb{I}}}}
\put(194,719){\makebox(0,0)[b]{\raisebox{0pt}[0pt][0pt]{\elvrm V}}}
\put(291,708){\makebox(0,0)[b]{\raisebox{0pt}[0pt][0pt]{\elvrm \xb{C}}}}
\put(328,692){\makebox(0,0)[b]{\raisebox{0pt}[0pt][0pt]{\elvrm \xp{I}}}}
\put(304,672){\makebox(0,0)[b]{\raisebox{0pt}[0pt][0pt]{\elvrm \xp{N}}}}
\put(268,672){\makebox(0,0)[b]{\raisebox{0pt}[0pt][0pt]{\ninrm $\emptyset$}}}
\put(304,655){\makebox(0,0)[b]{\raisebox{0pt}[0pt][0pt]{\ninrm Alice}}}
\put(364,677){\makebox(0,0)[b]{\raisebox{0pt}[0pt][0pt]{\elvrm \xb{I}}}}
\put(341,640){\makebox(0,0)[b]{\raisebox{0pt}[0pt][0pt]{\ninrm will}}}
\put(194,698){\makebox(0,0)[b]{\raisebox{0pt}[0pt][0pt]{\ninrm think}}}
\put(213,739){\makebox(0,0)[b]{\raisebox{0pt}[0pt][0pt]{\elvrm \xb{V}}}}
\put(268,691){\makebox(0,0)[b]{\raisebox{0pt}[0pt][0pt]{\elvrm C}}}
\put(378,628){\makebox(0,0)[b]{\raisebox{0pt}[0pt][0pt]{\elvrm V}}}
\put( 43,798){\makebox(0,0)[b]{\raisebox{0pt}[0pt][0pt]{\elvrm \xp{N}\sub{i}}}}
\put(144,785){\makebox(0,0)[b]{\raisebox{0pt}[0pt][0pt]{\elvrm \xp{I}}}}
\put(213,753){\makebox(0,0)[b]{\raisebox{0pt}[0pt][0pt]{\elvrm \xp{V}}}}
\put(254,722){\makebox(0,0)[b]{\raisebox{0pt}[0pt][0pt]{\elvrm \xp{C}}}}
\end{picture}
\end{center}
\caption{Extraction from the object, S-Structure.\label{fig.sent.1s}}
\end{figure}
Figure~\ref{fig.sent.1s} gives the S-Structure of
a more or less typical GB analysis of the
sentence:
\enumsentence{Whom do you think Alice will invite.}
\begin{figure}
\begin{center}
\begin{picture}(0,0)%
\includegraphics{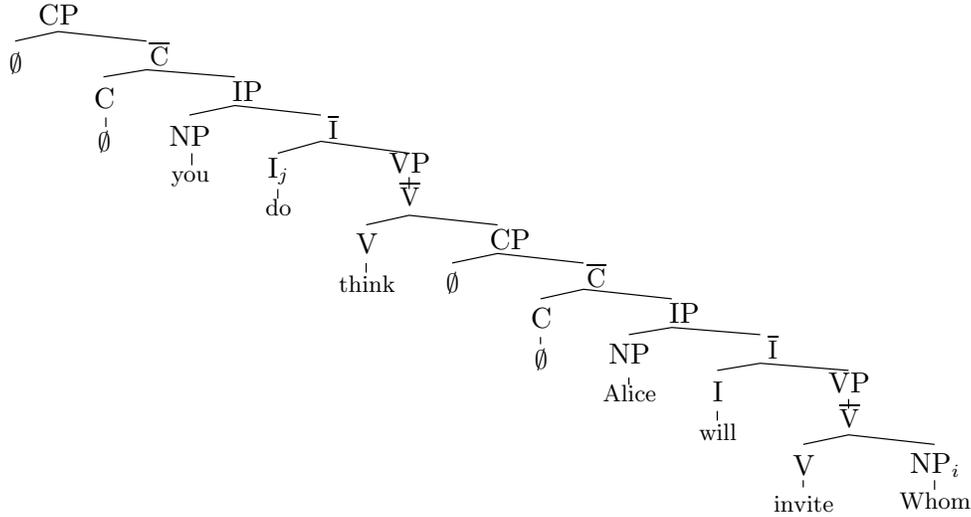}%
\end{picture}%
\setlength{\unitlength}{0.012500in}%
\begin{picture}(442,215)(14,619)
\put(148,759){\makebox(0,0)[b]{\raisebox{0pt}[0pt][0pt]{\elvrm I\sub{j}}}}
\put(148,743){\makebox(0,0)[b]{\raisebox{0pt}[0pt][0pt]{\ninrm do}}}
\put( 56,823){\makebox(0,0)[b]{\raisebox{0pt}[0pt][0pt]{\elvrm \xp{C}}}}
\put(203,747){\makebox(0,0)[b]{\raisebox{0pt}[0pt][0pt]{\elvrm \xb{V}}}}
\put(387,654){\makebox(0,0)[b]{\raisebox{0pt}[0pt][0pt]{\elvrm \xb{V}}}}
\put(203,761){\makebox(0,0)[b]{\raisebox{0pt}[0pt][0pt]{\elvrm \xp{V}}}}
\put(387,669){\makebox(0,0)[b]{\raisebox{0pt}[0pt][0pt]{\elvrm \xp{V}}}}
\put(423,635){\makebox(0,0)[b]{\raisebox{0pt}[0pt][0pt]{\elvrm \xp{N}\sub{i}}}}
\put(423,620){\makebox(0,0)[b]{\raisebox{0pt}[0pt][0pt]{\ninrm Whom}}}
\put(368,619){\makebox(0,0)[b]{\raisebox{0pt}[0pt][0pt]{\ninrm invite}}}
\put(368,634){\makebox(0,0)[b]{\raisebox{0pt}[0pt][0pt]{\elvrm V}}}
\put(332,649){\makebox(0,0)[b]{\raisebox{0pt}[0pt][0pt]{\ninrm will}}}
\put(332,665){\makebox(0,0)[b]{\raisebox{0pt}[0pt][0pt]{\elvrm I}}}
\put(295,681){\makebox(0,0)[b]{\raisebox{0pt}[0pt][0pt]{\elvrm \xp{N}}}}
\put(295,665){\makebox(0,0)[b]{\raisebox{0pt}[0pt][0pt]{\ninrm Alice}}}
\put(258,679){\makebox(0,0)[b]{\raisebox{0pt}[0pt][0pt]{\ninrm $\emptyset$}}}
\put(258,696){\makebox(0,0)[b]{\raisebox{0pt}[0pt][0pt]{\elvrm C}}}
\put(221,711){\makebox(0,0)[b]{\raisebox{0pt}[0pt][0pt]{\ninrm $\emptyset$}}}
\put(185,711){\makebox(0,0)[b]{\raisebox{0pt}[0pt][0pt]{\ninrm think}}}
\put(185,727){\makebox(0,0)[b]{\raisebox{0pt}[0pt][0pt]{\elvrm V}}}
\put(111,757){\makebox(0,0)[b]{\raisebox{0pt}[0pt][0pt]{\ninrm you}}}
\put(111,772){\makebox(0,0)[b]{\raisebox{0pt}[0pt][0pt]{\elvrm \xp{N}}}}
\put( 75,788){\makebox(0,0)[b]{\raisebox{0pt}[0pt][0pt]{\elvrm C}}}
\put( 38,803){\makebox(0,0)[b]{\raisebox{0pt}[0pt][0pt]{\ninrm $\emptyset$}}}
\put( 75,770){\makebox(0,0)[b]{\raisebox{0pt}[0pt][0pt]{\ninrm $\emptyset$}}}
\put( 98,806){\makebox(0,0)[b]{\raisebox{0pt}[0pt][0pt]{\elvrm \xb{C}}}}
\put(135,791){\makebox(0,0)[b]{\raisebox{0pt}[0pt][0pt]{\elvrm \xp{I}}}}
\put(171,775){\makebox(0,0)[b]{\raisebox{0pt}[0pt][0pt]{\elvrm \xb{I}}}}
\put(245,729){\makebox(0,0)[b]{\raisebox{0pt}[0pt][0pt]{\elvrm \xp{C}}}}
\put(281,713){\makebox(0,0)[b]{\raisebox{0pt}[0pt][0pt]{\elvrm \xb{C}}}}
\put(318,698){\makebox(0,0)[b]{\raisebox{0pt}[0pt][0pt]{\elvrm \xp{I}}}}
\put(355,683){\makebox(0,0)[b]{\raisebox{0pt}[0pt][0pt]{\elvrm \xb{I}}}}
\end{picture}
\end{center}
\caption{Extraction from the object, D-Structure.\label{fig.sent.1d}}
\end{figure}
In the D-Structure (Figure~\ref{fig.sent.1d}) the element carrying the
inflection is positioned between the subject and the predicate and {\em
Whom\/} is in its standard position as the object of {\em invite\/}.
Move-$\alpha$ transforms this structure by cutting out the subtrees rooted at
I\sub{j} and \xp{N}\sub{i}, leaving phonetically empty traces (t\sub{j} and
t\sub{i}), and re-attaching them a higher positions in the tree.
In the case of {\em Whom\/} the movement occurs in two steps, with
traces being left at each intermediate position.  The original position
of the moved element is referred to as the {\em base\/} position, and its
final resting place is the {\em target\/} position.  The moved element is
identified with its traces by co-indexation.  Together, an element and
the traces co-indexed with it form a {\em chain\/}.  Chains can be broken
up into a sequence of {\em links\/} each consisting of a trace and its
{\em antecedent\/}---the next higher element of the chain.

The fundamental issue we must address in defining chains within $\LKP$
is how to identify the antecedent of a trace without reference to
indices.  Our key idea is that, if we can limit the portion of the tree
in which an antecedent can occur, then we can possibly bound the number
of potential antecedents a trace may have.  Such a bound would suffice
since, while we cannot capture indexations with an unbounded range of
index, we can capture any indexation in which there is a constant bound on 
the total number of distinct indices.

In the standard GB account of movement, that of Barriers~\cite{chomsk86},
there are two principles that tend to bound the length of links.  The
first is {\em $n$-subjacency\/}, which, roughly, limits the number of
phrasal boundaries that a link can cross.  This is exactly the kind of
constraint we need.  Unfortunately it is responsible only for weak
effects; there are many sentences that violate $n$-subjacency that
are only of degraded acceptability rather than outright ungrammatical.
The second principle that might do is the {\em Empty Category
Principle\/}.  This puts specific constraints on the structural
relationship between a trace and its antecedent.  Indices, however, play a
significant role in Chomsky's formulation of this principle.

There is a formulation of ECP, due to Rizzi and
based on his notion of {\em Relativized Minimality\/}~\cite{rizzi90}, in which
the role of indexation is largely eliminated.  In Rizzi's theory,
this is a conjunctive principle with two components, a Formal Licensing
requirement and an Identification requirement:
\begin{description}
\item[{\em ECP (Rizzi):}]\
\begin{itemize}
\item A non-pronominal empty category must be properly head-governed.
(Formal Licensing)
\item Operators must be identified with their variables. (Identification)
\end{itemize}
\end{description}
We are interested in the identification requirement, which, incidently,
is responsible for most of the effects attributed to ECP in the Barriers
account.  This constraint requires every trace (variable) to be identified with
its target (operator).  This can be done in one of two ways,  either by 
a particular class of index, the referential indices,
or by a sequence of {\em antecedent-government\/}
links.
In the latter case the role of indices in identifying chains can be
taken over by the antecedent-government relation.

To a first approximation, government is simply a relation between an
element and those elements occurring in a specifically limited region of
the tree dominated by the phrase in which that element (the governor)
occurs.  Its definition
 has three components.  First, for the class of government
relations we are considering here, the governor must c-command the
elements it governs, that is, those elements must be dominated by a sibling 
of the governor.  Second, there must be no intervening barrier.  For
Rizzi, the notion of barrier is much weaker than it is in the Barriers
account.  Here, this constraint simply forbids the government relation
from crossing certain phrasal boundaries (in particular specifiers,
adjuncts and complements of nouns or prepositions).  The final component
of the government relation requires a governor to be the minimal
potential governor of the elements it governs, that is, no potential
governor can fall properly between a governor and the elements it
governs.  There are a range of types of government relations that fall under
this general category.  In Rizzi's theory 
only potential governors of the same type count for the minimality
requirement.  (This is the relativized aspect of his theory.)  For
antecedent-government there is an additional requirement that the
governor be co-indexed with the trace.
\begin{definition}
x {\em antecedent-governs} y iff
\begin{itemize}
\item x c-commands y.
\item No barrier falls between x and y.
\item Minimality is respected.
\item x and y are co-indexed.
\end{itemize}
\end{definition}
As we will see, we can drop the co-indexation
requirement on the grounds that, when it
exists, the antecedent-governor is unique.

\begin{figure}
\begin{center}
\begin{picture}(0,0)%
\includegraphics{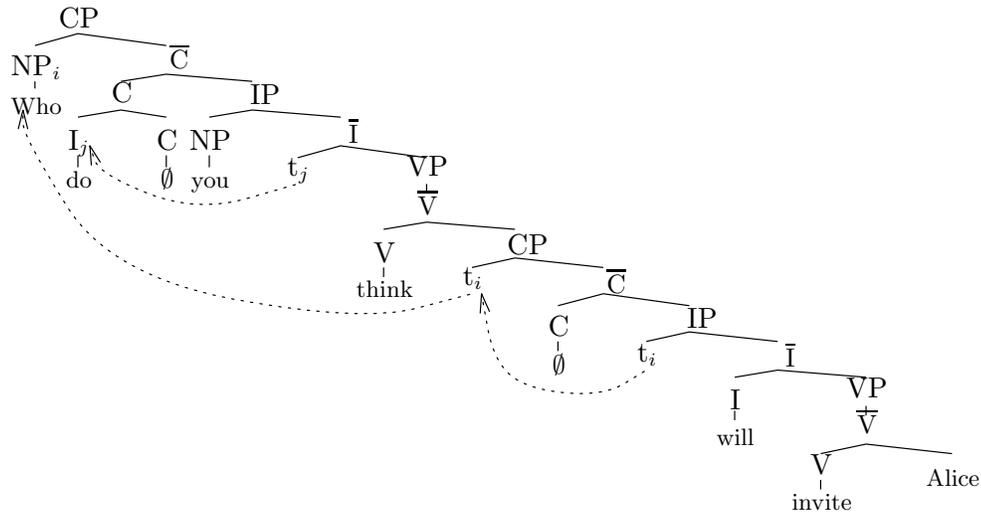}%
\end{picture}%
\setlength{\unitlength}{0.012500in}%
\begin{picture}(427,213)(15,615)
\put( 66,817){\makebox(0,0)[b]{\raisebox{0pt}[0pt][0pt]{\elvrm \xp{C}}}}
\put( 84,786){\makebox(0,0)[b]{\raisebox{0pt}[0pt][0pt]{\elvrm C}}}
\put(158,755){\makebox(0,0)[b]{\raisebox{0pt}[0pt][0pt]{\elvrm t\sub{j}}}}
\put(212,755){\makebox(0,0)[b]{\raisebox{0pt}[0pt][0pt]{\elvrm \xp{V}}}}
\put(212,739){\makebox(0,0)[b]{\raisebox{0pt}[0pt][0pt]{\elvrm \xb{V}}}}
\put(231,709){\makebox(0,0)[b]{\raisebox{0pt}[0pt][0pt]{\elvrm t\sub{i}}}}
\put(396,662){\makebox(0,0)[b]{\raisebox{0pt}[0pt][0pt]{\elvrm \xp{V}}}}
\put(396,647){\makebox(0,0)[b]{\raisebox{0pt}[0pt][0pt]{\elvrm \xb{V}}}}
\put(377,630){\makebox(0,0)[b]{\raisebox{0pt}[0pt][0pt]{\elvrm V}}}
\put(377,615){\makebox(0,0)[b]{\raisebox{0pt}[0pt][0pt]{\ninrm invite}}}
\put(304,677){\makebox(0,0)[b]{\raisebox{0pt}[0pt][0pt]{\elvrm t\sub{i}}}}
\put( 48,781){\makebox(0,0)[b]{\raisebox{0pt}[0pt][0pt]{\ninrm Who}}}
\put( 48,797){\makebox(0,0)[b]{\raisebox{0pt}[0pt][0pt]{\elvrm \xp{N}\sub{i}}}}
\put( 66,750){\makebox(0,0)[b]{\raisebox{0pt}[0pt][0pt]{\ninrm do}}}
\put( 66,766){\makebox(0,0)[b]{\raisebox{0pt}[0pt][0pt]{\elvrm I\sub{j}}}}
\put(103,750){\makebox(0,0)[b]{\raisebox{0pt}[0pt][0pt]{\ninrm $\emptyset$}}}
\put(103,766){\makebox(0,0)[b]{\raisebox{0pt}[0pt][0pt]{\elvrm C}}}
\put(121,750){\makebox(0,0)[b]{\raisebox{0pt}[0pt][0pt]{\ninrm you}}}
\put(121,766){\makebox(0,0)[b]{\raisebox{0pt}[0pt][0pt]{\elvrm \xp{N}}}}
\put(194,704){\makebox(0,0)[b]{\raisebox{0pt}[0pt][0pt]{\ninrm think}}}
\put(194,719){\makebox(0,0)[b]{\raisebox{0pt}[0pt][0pt]{\elvrm V}}}
\put(267,672){\makebox(0,0)[b]{\raisebox{0pt}[0pt][0pt]{\ninrm $\emptyset$}}}
\put(267,688){\makebox(0,0)[b]{\raisebox{0pt}[0pt][0pt]{\elvrm C}}}
\put(341,642){\makebox(0,0)[b]{\raisebox{0pt}[0pt][0pt]{\ninrm will}}}
\put(341,657){\makebox(0,0)[b]{\raisebox{0pt}[0pt][0pt]{\elvrm I}}}
\put(432,625){\makebox(0,0)[b]{\raisebox{0pt}[0pt][0pt]{\ninrm Alice}}}
\put(108,800){\makebox(0,0)[b]{\raisebox{0pt}[0pt][0pt]{\elvrm \xb{C}}}}
\put(144,785){\makebox(0,0)[b]{\raisebox{0pt}[0pt][0pt]{\elvrm \xp{I}}}}
\put(181,769){\makebox(0,0)[b]{\raisebox{0pt}[0pt][0pt]{\elvrm \xb{I}}}}
\put(254,722){\makebox(0,0)[b]{\raisebox{0pt}[0pt][0pt]{\elvrm \xp{C}}}}
\put(291,706){\makebox(0,0)[b]{\raisebox{0pt}[0pt][0pt]{\elvrm \xb{C}}}}
\put(327,691){\makebox(0,0)[b]{\raisebox{0pt}[0pt][0pt]{\elvrm \xp{I}}}}
\put(364,675){\makebox(0,0)[b]{\raisebox{0pt}[0pt][0pt]{\elvrm \xb{I}}}}
\end{picture}
\end{center}
\caption{Extraction from the subject.\label{fig.sent.2}}
\end{figure}
As an example of these relationships, consider, in Figure~\ref{fig.sent.2}, the
trace in the {\em specifier\/} of the lower \xp{I}, that is, the trace of {\em
Who\/} falling immediately under the \xp{I}.  The elements
c-commanding this trace include the (empty) {C}, the t\sub{i}
in the specifier of \xp{C}, the {V}, etc.  This is a {\em Wh-Trace\/}
which means that, by the principles of Binding Theory,
its antecedent must fall in a {\em non-argument\/}
position.  In the example, the non-argument positions c-commanding the
trace are just the specifiers of the \xp{C}s.  By minimality, no
potential antecedent of the trace beyond the closest specifier of
\xp{C} can govern it.  Thus the only possible antecedent-governor
of the trace in question is the trace in the specifier of
the lower \xp{C}, which is, in fact, its antecedent.

\begin{figure}
\begin{center}
\begin{picture}(0,0)%
\includegraphics{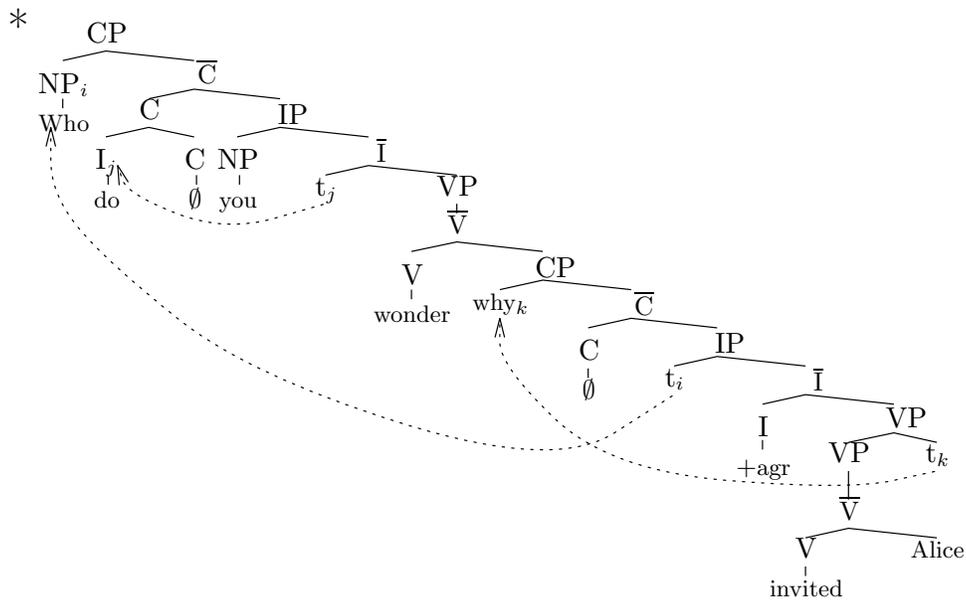}%
\end{picture}%
\setlength{\unitlength}{0.012500in}%
\begin{picture}(418,243)(15,585)
\put( 66,817){\makebox(0,0)[b]{\raisebox{0pt}[0pt][0pt]{\elvrm \xp{C}}}}
\put( 84,785){\makebox(0,0)[b]{\raisebox{0pt}[0pt][0pt]{\elvrm C}}}
\put(158,753){\makebox(0,0)[b]{\raisebox{0pt}[0pt][0pt]{\elvrm t\sub{j}}}}
\put(213,753){\makebox(0,0)[b]{\raisebox{0pt}[0pt][0pt]{\elvrm \xp{V}}}}
\put(213,737){\makebox(0,0)[b]{\raisebox{0pt}[0pt][0pt]{\elvrm \xb{V}}}}
\put(377,641){\makebox(0,0)[b]{\raisebox{0pt}[0pt][0pt]{\elvrm \xp{V}}}}
\put(414,641){\makebox(0,0)[b]{\raisebox{0pt}[0pt][0pt]{\elvrm t\sub{k}}}}
\put(377,617){\makebox(0,0)[b]{\raisebox{0pt}[0pt][0pt]{\elvrm \xb{V}}}}
\put(359,601){\makebox(0,0)[b]{\raisebox{0pt}[0pt][0pt]{\elvrm V}}}
\put(359,585){\makebox(0,0)[b]{\raisebox{0pt}[0pt][0pt]{\ninrm invited}}}
\put(414,601){\makebox(0,0)[b]{\raisebox{0pt}[0pt][0pt]{\ninrm Alice}}}
\put(304,673){\makebox(0,0)[b]{\raisebox{0pt}[0pt][0pt]{\elvrm t\sub{i}}}}
\put( 29,817){\makebox(0,0)[b]{\raisebox{0pt}[0pt][0pt]{\svtnrm *}}}
\put( 48,780){\makebox(0,0)[b]{\raisebox{0pt}[0pt][0pt]{\ninrm Who}}}
\put( 48,796){\makebox(0,0)[b]{\raisebox{0pt}[0pt][0pt]{\elvrm \xp{N}\sub{i}}}}
\put( 66,748){\makebox(0,0)[b]{\raisebox{0pt}[0pt][0pt]{\ninrm do}}}
\put( 66,764){\makebox(0,0)[b]{\raisebox{0pt}[0pt][0pt]{\elvrm I\sub{j}}}}
\put(103,764){\makebox(0,0)[b]{\raisebox{0pt}[0pt][0pt]{\elvrm C}}}
\put(103,748){\makebox(0,0)[b]{\raisebox{0pt}[0pt][0pt]{\ninrm $\emptyset$}}}
\put(121,748){\makebox(0,0)[b]{\raisebox{0pt}[0pt][0pt]{\ninrm you}}}
\put(121,764){\makebox(0,0)[b]{\raisebox{0pt}[0pt][0pt]{\elvrm \xp{N}}}}
\put(194,700){\makebox(0,0)[b]{\raisebox{0pt}[0pt][0pt]{\ninrm wonder}}}
\put(194,716){\makebox(0,0)[b]{\raisebox{0pt}[0pt][0pt]{\elvrm V}}}
\put(268,668){\makebox(0,0)[b]{\raisebox{0pt}[0pt][0pt]{\ninrm $\emptyset$}}}
\put(268,684){\makebox(0,0)[b]{\raisebox{0pt}[0pt][0pt]{\elvrm C}}}
\put(341,636){\makebox(0,0)[b]{\raisebox{0pt}[0pt][0pt]{\ninrm $+$agr}}}
\put(341,652){\makebox(0,0)[b]{\raisebox{0pt}[0pt][0pt]{\elvrm I}}}
\put(108,800){\makebox(0,0)[b]{\raisebox{0pt}[0pt][0pt]{\elvrm \xb{C}}}}
\put(144,783){\makebox(0,0)[b]{\raisebox{0pt}[0pt][0pt]{\elvrm \xp{I}}}}
\put(181,767){\makebox(0,0)[b]{\raisebox{0pt}[0pt][0pt]{\elvrm \xb{I}}}}
\put(254,719){\makebox(0,0)[b]{\raisebox{0pt}[0pt][0pt]{\elvrm \xp{C}}}}
\put(291,703){\makebox(0,0)[b]{\raisebox{0pt}[0pt][0pt]{\elvrm \xb{C}}}}
\put(327,687){\makebox(0,0)[b]{\raisebox{0pt}[0pt][0pt]{\elvrm \xp{I}}}}
\put(364,671){\makebox(0,0)[b]{\raisebox{0pt}[0pt][0pt]{\elvrm \xb{I}}}}
\put(401,656){\makebox(0,0)[b]{\raisebox{0pt}[0pt][0pt]{\elvrm \xp{V}}}}
\put(231,705){\makebox(0,0)[b]{\raisebox{0pt}[0pt][0pt]{\ninrm why\sub{k}}}}
\end{picture}
\end{center}
\caption{An ECP violation.\label{fig.sent.3}}
\end{figure}
In contrast, if we fill that position with a moved adverbial, as in the
example of Figure~\ref{fig.sent.3}, there is a problem.  The element
{\em why\/} cannot be the antecedent of the trace in the specifier of the
lower \xp{I}, but it blocks government by all other potential
antecedents.  Thus the trace t\sub{i} cannot be identified with its
antecedent, and the sentence is ruled ungrammatical on the grounds that
it violates ECP.

In this way, minimality suffices to pick out the unique antecedent of
traces in chains that are identified by antecedent-government.  But
under Rizzi's criteria chains can also be identified by
referential indices.  These are just indices assigned to elements that
receive what are termed {\em referential\/} Theta roles.  Again to a
first approximation, we can take these simply to be elements that are
the objects of verbs.  
\begin{figure}
\begin{center}
\begin{picture}(0,0)%
\includegraphics{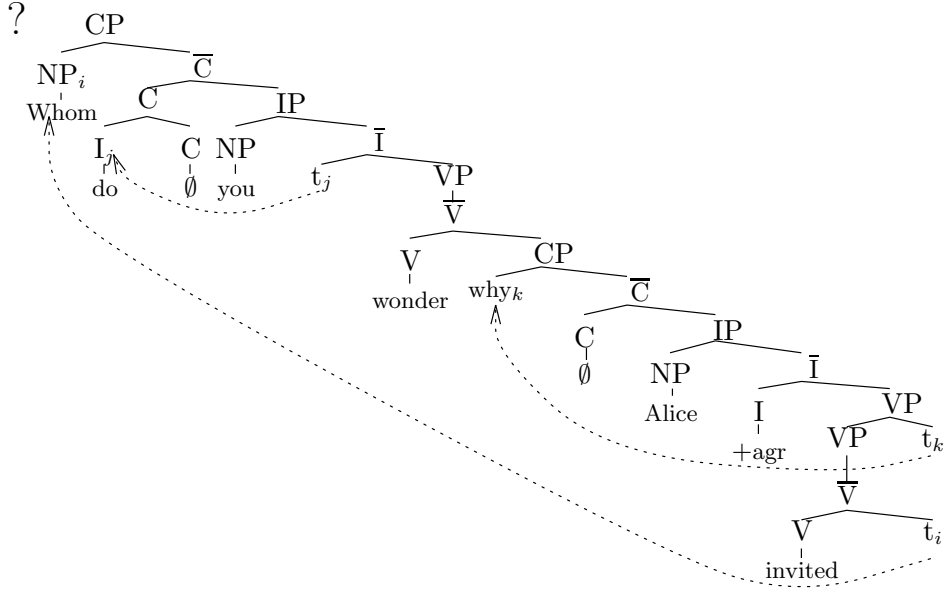}%
\end{picture}%
\setlength{\unitlength}{0.012500in}%
\begin{picture}(417,244)(15,585)
\put( 66,817){\makebox(0,0)[b]{\raisebox{0pt}[0pt][0pt]{\elvrm \xp{C}}}}
\put( 84,786){\makebox(0,0)[b]{\raisebox{0pt}[0pt][0pt]{\elvrm C}}}
\put(157,754){\makebox(0,0)[b]{\raisebox{0pt}[0pt][0pt]{\elvrm t\sub{j}}}}
\put(212,754){\makebox(0,0)[b]{\raisebox{0pt}[0pt][0pt]{\elvrm \xp{V}}}}
\put(212,738){\makebox(0,0)[b]{\raisebox{0pt}[0pt][0pt]{\elvrm \xb{V}}}}
\put(377,644){\makebox(0,0)[b]{\raisebox{0pt}[0pt][0pt]{\elvrm \xp{V}}}}
\put(413,644){\makebox(0,0)[b]{\raisebox{0pt}[0pt][0pt]{\elvrm t\sub{k}}}}
\put(377,620){\makebox(0,0)[b]{\raisebox{0pt}[0pt][0pt]{\elvrm \xb{V}}}}
\put(358,605){\makebox(0,0)[b]{\raisebox{0pt}[0pt][0pt]{\elvrm V}}}
\put(413,605){\makebox(0,0)[b]{\raisebox{0pt}[0pt][0pt]{\elvrm t\sub{i}}}}
\put(358,589){\makebox(0,0)[b]{\raisebox{0pt}[0pt][0pt]{\ninrm invited}}}
\put( 29,817){\makebox(0,0)[b]{\raisebox{0pt}[0pt][0pt]{\svtnrm ?}}}
\put( 48,781){\makebox(0,0)[b]{\raisebox{0pt}[0pt][0pt]{\ninrm Whom}}}
\put( 48,796){\makebox(0,0)[b]{\raisebox{0pt}[0pt][0pt]{\elvrm \xp{N}\sub{i}}}}
\put( 66,749){\makebox(0,0)[b]{\raisebox{0pt}[0pt][0pt]{\ninrm do}}}
\put( 66,765){\makebox(0,0)[b]{\raisebox{0pt}[0pt][0pt]{\elvrm I\sub{j}}}}
\put(102,749){\makebox(0,0)[b]{\raisebox{0pt}[0pt][0pt]{\ninrm $\emptyset$}}}
\put(102,765){\makebox(0,0)[b]{\raisebox{0pt}[0pt][0pt]{\elvrm C}}}
\put(121,749){\makebox(0,0)[b]{\raisebox{0pt}[0pt][0pt]{\ninrm you}}}
\put(121,765){\makebox(0,0)[b]{\raisebox{0pt}[0pt][0pt]{\elvrm \xp{N}}}}
\put(194,702){\makebox(0,0)[b]{\raisebox{0pt}[0pt][0pt]{\ninrm wonder}}}
\put(194,718){\makebox(0,0)[b]{\raisebox{0pt}[0pt][0pt]{\elvrm V}}}
\put(267,671){\makebox(0,0)[b]{\raisebox{0pt}[0pt][0pt]{\ninrm $\emptyset$}}}
\put(267,686){\makebox(0,0)[b]{\raisebox{0pt}[0pt][0pt]{\elvrm C}}}
\put(303,655){\makebox(0,0)[b]{\raisebox{0pt}[0pt][0pt]{\ninrm Alice}}}
\put(303,671){\makebox(0,0)[b]{\raisebox{0pt}[0pt][0pt]{\elvrm \xp{N}}}}
\put(340,639){\makebox(0,0)[b]{\raisebox{0pt}[0pt][0pt]{\ninrm $+$agr}}}
\put(340,655){\makebox(0,0)[b]{\raisebox{0pt}[0pt][0pt]{\elvrm I}}}
\put(107,799){\makebox(0,0)[b]{\raisebox{0pt}[0pt][0pt]{\elvrm \xb{C}}}}
\put(144,784){\makebox(0,0)[b]{\raisebox{0pt}[0pt][0pt]{\elvrm \xp{I}}}}
\put(181,768){\makebox(0,0)[b]{\raisebox{0pt}[0pt][0pt]{\elvrm \xb{I}}}}
\put(254,721){\makebox(0,0)[b]{\raisebox{0pt}[0pt][0pt]{\elvrm \xp{C}}}}
\put(290,705){\makebox(0,0)[b]{\raisebox{0pt}[0pt][0pt]{\elvrm \xb{C}}}}
\put(327,689){\makebox(0,0)[b]{\raisebox{0pt}[0pt][0pt]{\elvrm \xp{I}}}}
\put(363,673){\makebox(0,0)[b]{\raisebox{0pt}[0pt][0pt]{\elvrm \xb{I}}}}
\put(400,658){\makebox(0,0)[b]{\raisebox{0pt}[0pt][0pt]{\elvrm \xp{V}}}}
\put(230,707){\makebox(0,0)[b]{\raisebox{0pt}[0pt][0pt]{\ninrm why\sub{k}}}}
\end{picture}
\end{center}
\caption{A 1-subjacency violation.\label{fig.sent.4}}
\end{figure}
In Figure~\ref{fig.sent.3} {\em Who\/} is extracted from
the embedded subject.
If we return to our original example, in which we
extract from the object, we find that filling the specifier of the lower
\xp{C} with a moved adverbial (Figure~\ref{fig.sent.4}) has a less
dramatic effect.  While antecedent government of the trace in the
complement of the lower \xp{V} is  blocked, that trace can now be
identified with its target by the referential index they share.  The
fact that this example is not judged to be as bad as the example from
Figure~\ref{fig.sent.3} is attributed, then, to the fact that it is only
a 1-subjacency violation rather than an ECP violation.

In general, we could be forced to resort to a mechanism equivalent to
indexation in order to distinguish such referential chains.  It turns out,
however, that in English, at least, chains of this type do not overlap.
Manzini~\cite{manzini92}, in fact, argues for an account of \xb{A}-movement
(movements, like these we have been considering, to non-argument positions)
which implies that no more than two such chains---one 
referential and one non-referential---may ever overlap.
Thus, we need to identify only a single referential antecedent in any
single context.

\subsection{Defining Antecedent-Government, Links, and Chains}
{\em Relativized Minimality} theory distinguishes a number of distinct
varieties of antecedent-government, one for each class of movement.  
We look at one representative case
\Abar-antecedent-government.  This is defined, in $\LKP$ as follows:
\begin{eqnarray*}
\lefteqn{\f{\Abar-Antecedent-Governs}(x,y) \equiv}
\\
&&
\neg\f{A-pos}(x)\land\f{C-Commands}(x,y)\land\f{T.Eq}(x,y)\land
\\
&&\bcomment{---$x$ is a potential antecedent in an \Abar-position}\\
&&
\neg(\exists z)[\f{Intervening-Barrier}(z,x,y)]\land\\
&&\bcomment{---no barrier intervenes}\\
&&
\neg(\exists z)[\f{Spec}(z)\land\neg\f{A-pos}(z)\land\\
&&
\hphantom{\neg(\exists z)[}\f{C-Commands}(z,x)\land
\f{Intervenes}(z,x,y)]\\
&&\bcomment{---minimality is respected}\\
\end{eqnarray*}
In words, this says simply that $x$ is an \Abar-antecedent-governor of $y$ iff
$x$ is in a non-argument (\Abar) position, it c-commands $y$, no barrier
intervenes between $x$ and $y$, and no non-argument specifier falls between $x$
and $y$.  The actual definitions of $\f{A-Pos}$, $\f{T.Eq}$,
$\f{Intervening-Barrier}$, $\f{Spec}$, and $\f{Intervenes}$ is unimportant
here.   The predicate $\f{T.Eq}$ is used to check the compatibility of the
features of the trace with those of its antecedent.

Using this, we can define the link relation. 
\begin{eqnarray*}
\lefteqn{\f{\Abar-\Rbar-Link}(x,y) \equiv}\\
&&\f{\Abar-Antecedent-Governs}(x,y)\land\neg\f{Ref}(x)\land\neg\f{Ref}(y)
\land\\
&&
\f{Bar2}(x)\land(\neg\f{Target}(x)\lor\f{Spec}(x))\land\\
&&\bcomment{---$x$ is an \xp{X} and is a specifier if it is the target}
\\
&&
\neg\f{Base}(x)\land\f{Trace}(y)\land
\f{$-$anaphor}(y)\land\f{$-$pronominal}(y)\\
&&\bcomment{---$y$ is an \Abar-trace, $x$ is not in Base position}\\
\end{eqnarray*}
This is just antecedent-government with certain additional configurational
requirements.  We can extend the notion of
links based on Rizzi's antecedent-government to include antecedents and
traces that Rizzi  identifies with a referential index
(which we refer to as \Abar-referential links), and links formed by
rightward movement.  This gives us five distinct link relations.  As they
are mutually exclusive, we can take their disjunction to form a single link
relation which must be satisfied by every trace and its antecedent.
\begin{eqnarray*}
\f{Link}(x,y) &\equiv&\f{A-Link}(x,y)\lor\f{\Abar-\Rbar-Link}(x,y)\lor
\\
&&\f{\Abar-Ref-Link}(x,y)\lor\f{\xz{X}-Link}(x,y)\lor\\
&&\f{Right-Link}(x,y)
\end{eqnarray*}

The idea, now, is to define chains as any set of nodes that are
linearly ordered by $\f{Link}$.  Before we can do this, though, we have one
more issue to resolve.  The problem is that, while we can identify a unique
antecedent for each trace, nothing assures us that there will be a unique trace
for each antecedent, that is, nothing prevents us from identifying the same
node as the antecedent of more than one trace.
\begin{figure}
\begin{center}
\begin{picture}(0,0)%
\includegraphics{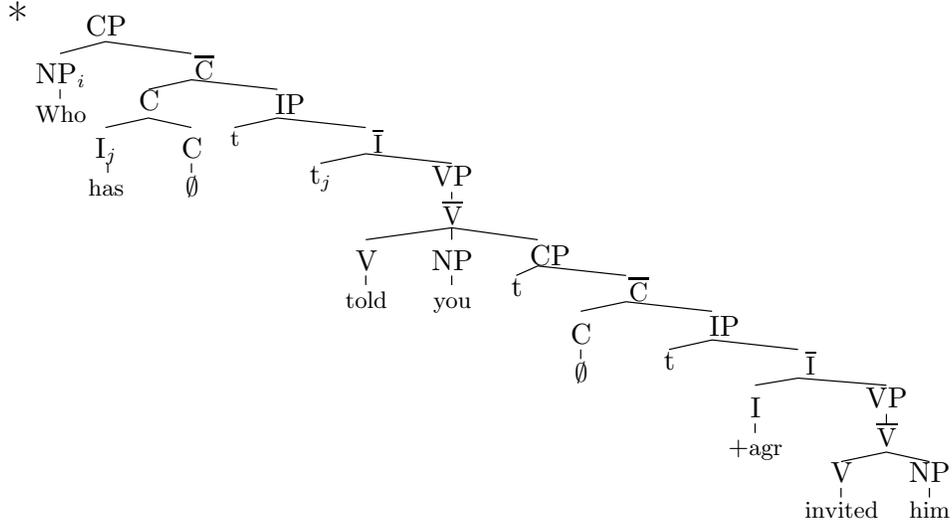}%
\end{picture}%
\setlength{\unitlength}{0.012500in}%
\begin{picture}(413,213)(14,615)
\put( 84,786){\makebox(0,0)[b]{\raisebox{0pt}[0pt][0pt]{\elvrm C}}}
\put(156,755){\makebox(0,0)[b]{\raisebox{0pt}[0pt][0pt]{\elvrm t\sub{j}}}}
\put(238,708){\makebox(0,0)[b]{\raisebox{0pt}[0pt][0pt]{\elvrm t}}}
\put(374,630){\makebox(0,0)[b]{\raisebox{0pt}[0pt][0pt]{\elvrm V}}}
\put(374,615){\makebox(0,0)[b]{\raisebox{0pt}[0pt][0pt]{\ninrm invited}}}
\put(411,630){\makebox(0,0)[b]{\raisebox{0pt}[0pt][0pt]{\elvrm \xp{N}}}}
\put(411,615){\makebox(0,0)[b]{\raisebox{0pt}[0pt][0pt]{\ninrm him}}}
\put( 66,817){\makebox(0,0)[b]{\raisebox{0pt}[0pt][0pt]{\elvrm \xp{C}}}}
\put(120,771){\makebox(0,0)[b]{\raisebox{0pt}[0pt][0pt]{\ninrm t}}}
\put(302,677){\makebox(0,0)[b]{\raisebox{0pt}[0pt][0pt]{\elvrm t}}}
\put( 29,817){\makebox(0,0)[b]{\raisebox{0pt}[0pt][0pt]{\svtnrm *}}}
\put( 47,781){\makebox(0,0)[b]{\raisebox{0pt}[0pt][0pt]{\ninrm Who}}}
\put( 47,797){\makebox(0,0)[b]{\raisebox{0pt}[0pt][0pt]{\elvrm \xp{N}\sub{i}}}}
\put( 66,750){\makebox(0,0)[b]{\raisebox{0pt}[0pt][0pt]{\ninrm has}}}
\put( 66,766){\makebox(0,0)[b]{\raisebox{0pt}[0pt][0pt]{\elvrm I\sub{j}}}}
\put(102,750){\makebox(0,0)[b]{\raisebox{0pt}[0pt][0pt]{\ninrm $\emptyset$}}}
\put(102,766){\makebox(0,0)[b]{\raisebox{0pt}[0pt][0pt]{\elvrm C}}}
\put(175,703){\makebox(0,0)[b]{\raisebox{0pt}[0pt][0pt]{\ninrm told}}}
\put(175,719){\makebox(0,0)[b]{\raisebox{0pt}[0pt][0pt]{\elvrm V}}}
\put(211,703){\makebox(0,0)[b]{\raisebox{0pt}[0pt][0pt]{\ninrm you}}}
\put(211,719){\makebox(0,0)[b]{\raisebox{0pt}[0pt][0pt]{\elvrm \xp{N}}}}
\put(265,672){\makebox(0,0)[b]{\raisebox{0pt}[0pt][0pt]{\ninrm $\emptyset$}}}
\put(265,688){\makebox(0,0)[b]{\raisebox{0pt}[0pt][0pt]{\elvrm C}}}
\put(338,641){\makebox(0,0)[b]{\raisebox{0pt}[0pt][0pt]{\ninrm $+$agr}}}
\put(338,657){\makebox(0,0)[b]{\raisebox{0pt}[0pt][0pt]{\elvrm I}}}
\put(107,799){\makebox(0,0)[b]{\raisebox{0pt}[0pt][0pt]{\elvrm \xb{C}}}}
\put(143,784){\makebox(0,0)[b]{\raisebox{0pt}[0pt][0pt]{\elvrm \xp{I}}}}
\put(180,768){\makebox(0,0)[b]{\raisebox{0pt}[0pt][0pt]{\elvrm \xb{I}}}}
\put(252,721){\makebox(0,0)[b]{\raisebox{0pt}[0pt][0pt]{\elvrm \xp{C}}}}
\put(289,706){\makebox(0,0)[b]{\raisebox{0pt}[0pt][0pt]{\elvrm \xb{C}}}}
\put(325,691){\makebox(0,0)[b]{\raisebox{0pt}[0pt][0pt]{\elvrm \xp{I}}}}
\put(361,675){\makebox(0,0)[b]{\raisebox{0pt}[0pt][0pt]{\elvrm \xb{I}}}}
\put(393,645){\makebox(0,0)[b]{\raisebox{0pt}[0pt][0pt]{\elvrm \xb{V}}}}
\put(393,661){\makebox(0,0)[b]{\raisebox{0pt}[0pt][0pt]{\elvrm \xp{V}}}}
\put(211,738){\makebox(0,0)[b]{\raisebox{0pt}[0pt][0pt]{\elvrm \xb{V}}}}
\put(211,754){\makebox(0,0)[b]{\raisebox{0pt}[0pt][0pt]{\elvrm \xp{V}}}}
\end{picture}
\end{center}
\caption{Conflated chains.\label{fig.conflate}}
\end{figure}
As an example, we might license
the tree in Figure~\ref{fig.conflate}.  This is the conflation of two
sentences: 
\eenumsentence{\item Who\sub{i} has t\sub{i} told you Alice invited him.
\item Who\sub{i} has Alice told you t\sub{i} t\sub{i} invited him.}
In the first we have extracted {\em Who\/} from the subject of the matrix
clause and in the second we have extracted it from the subject of the embedded
clause.  We can find a link relation between {\em Who\/} and the trace in the
specifier of the matrix \xp{I} and a link relation between {\em Who} and the
trace in the specifier of the embedded \xp{C}, but clearly it cannot have moved
from both positions. 

We rule out such structures by requiring that chains not only be linearly
ordered by $\f{Link}$, but that they are also closed under the link relation,
that is,  every chain includes every node that is related by $\f{Link}$ to any
node in the chain.  Trees like the the one in Figure~\ref{fig.conflate} are
ruled out on the grounds that any chain that contains either of the traces in
question must include both of them, and will therefore not be linearly ordered.

Formalizing this, we get:
\begin{eqnarray*}
\lefteqn{\f{Chain}(X) \equiv}\\
&&(\exists! x)[X(x)\land\f{Target}(x)]\land
(\exists! x)[X(x)\land\f{Base}(x)]\land\\
&&\bcomment{---$X$ contains exactly one Target and one Base}\\
&&
(\forall x)[X(x)\land\neg\f{Target}(x)\limp
  (\exists!y)[X(y)\land\f{Link}(y,x)]]\quad\land\\
&&\bcomment{---All non-Target have a unique antecedent in $X$}\\
&&
(\forall x)[X(x)\land\neg\f{Base}(x)\limp
  (\exists!y)[X(y)\land\f{Link}(x,y)]]\quad\land\\
&&\bcomment{---All non-Base have a unique successor in $X$}\\
&&
(\forall x,y)[X(x)\land(\f{Link}(x,y)\lor\f{Link}(y,x))\limp X(y)]
 \\
&&\bcomment{---$X$ is closed wrt the Link relation}.
\end{eqnarray*}

\subsection{Defining the ECP}
We can now capture Rizzi's version of the Empty Category Principle:

\noindent{\bf Licensing}
\[
(\forall x)[\f{Trace}(x)\limp
(\f{Bar0}(x)\lor(\exists y)[\f{Proper-Head-Governs}(y,x)])]
\]

\noindent{\bf Identification}
\[(\forall x)[\f{Trace}(x)\limp
(\exists X)[\f{Chain}(X)\land X(x)]]
\]

Note, in particular, that in our definition the identification requirement
is reduced simply to a requirement that every trace is a member of some
well-formed chain.  As we admit the notion of {\em trivial\/} chains---chains
with a single element, formed by zero movements---we can generalize this to a
global requirement that every element of the tree is a member of a (possibly
trivial) well-formed chain.

\noindent{\bf Identification (Generalized)}
\[
(\forall x)(\exists X)[\f{Chain}(X)\land X(x)].
\]

Recall that identification is the component of Rizzi's definition  that
accounts for most of the effects attributed to ECP in the Barrier's account of
movement.  Thus we have reduced a variety of effects to a single simple global
principle.  Of course we have paid for this with a complex
definition of chains, but much of this complexity lies in the definition
of antecedent-government and Rizzi argues, on
linguistic grounds, for essentially this definition in
any case.  It is satisfying that we can recover its added complexity in the
form of a greatly simplified ECP.

\subsection{Limits of the Definition}
The fact that we can exhibit a definition in $\LKP$ of the class of trees
licensed by a specific GB account of English provides a strong complexity
result for that class of trees---it is strongly context-free.  We don't, on the
other hand, expect this formalization to work for GB theories in general, and,
in particular we don't expect it to work for a GB account of Universal Grammar.
\begin{figure}
\begin{center}
\begin{picture}(0,0)%
\includegraphics{crossb.pstex}%
\end{picture}%
\setlength{\unitlength}{0.012500in}%
\begin{picture}(404,424)(51,410)
\put( 77,823){\makebox(0,0)[b]{\raisebox{0pt}[0pt][0pt]{\elvrm \xp{C}}}}
\put( 77,800){\makebox(0,0)[b]{\raisebox{0pt}[0pt][0pt]{\elvrm \xb{C}}}}
\put( 59,778){\makebox(0,0)[b]{\raisebox{0pt}[0pt][0pt]{\elvrm C}}}
\put(123,778){\makebox(0,0)[b]{\raisebox{0pt}[0pt][0pt]{\elvrm \xp{I}}}}
\put(105,754){\makebox(0,0)[b]{\raisebox{0pt}[0pt][0pt]{\elvrm \xp{N}}}}
\put(168,754){\makebox(0,0)[b]{\raisebox{0pt}[0pt][0pt]{\elvrm \xb{I}}}}
\put(150,731){\makebox(0,0)[b]{\raisebox{0pt}[0pt][0pt]{\elvrm \xp{V}}}}
\put(150,709){\makebox(0,0)[b]{\raisebox{0pt}[0pt][0pt]{\elvrm \xb{V}}}}
\put(132,686){\makebox(0,0)[b]{\raisebox{0pt}[0pt][0pt]{\elvrm \xp{I}}}}
\put(168,662){\makebox(0,0)[b]{\raisebox{0pt}[0pt][0pt]{\elvrm \xb{I}}}}
\put(150,640){\makebox(0,0)[b]{\raisebox{0pt}[0pt][0pt]{\elvrm \xp{V}}}}
\put(114,662){\makebox(0,0)[b]{\raisebox{0pt}[0pt][0pt]{\elvrm \xp{N}}}}
\put(150,433){\makebox(0,0)[b]{\raisebox{0pt}[0pt][0pt]{\elvrm \xb{V}}}}
\put( 59,754){\makebox(0,0)[b]{\raisebox{0pt}[0pt][0pt]{\elvit dat}}}
\put(105,731){\makebox(0,0)[b]{\raisebox{0pt}[0pt][0pt]{\elvit Jan}}}
\put(114,640){\makebox(0,0)[b]{\raisebox{0pt}[0pt][0pt]{\elvit Piet}}}
\put(114,548){\makebox(0,0)[b]{\raisebox{0pt}[0pt][0pt]{\elvit Marie}}}
\put(100,456){\makebox(0,0)[b]{\raisebox{0pt}[0pt][0pt]{\elvit die kinderen}}}
\put(150,524){\makebox(0,0)[b]{\raisebox{0pt}[0pt][0pt]{\elvrm \xb{V}}}}
\put(132,502){\makebox(0,0)[b]{\raisebox{0pt}[0pt][0pt]{\elvrm \xp{I}}}}
\put(168,479){\makebox(0,0)[b]{\raisebox{0pt}[0pt][0pt]{\elvrm \xb{I}}}}
\put(150,456){\makebox(0,0)[b]{\raisebox{0pt}[0pt][0pt]{\elvrm \xp{V}}}}
\put(114,479){\makebox(0,0)[b]{\raisebox{0pt}[0pt][0pt]{\elvrm \xp{N}}}}
\put(150,617){\makebox(0,0)[b]{\raisebox{0pt}[0pt][0pt]{\elvrm \xb{V}}}}
\put(132,593){\makebox(0,0)[b]{\raisebox{0pt}[0pt][0pt]{\elvrm \xp{I}}}}
\put(168,571){\makebox(0,0)[b]{\raisebox{0pt}[0pt][0pt]{\elvrm \xb{I}}}}
\put(150,548){\makebox(0,0)[b]{\raisebox{0pt}[0pt][0pt]{\elvrm \xp{V}}}}
\put(114,571){\makebox(0,0)[b]{\raisebox{0pt}[0pt][0pt]{\elvrm \xp{N}}}}
\put(150,410){\makebox(0,0)[b]{\raisebox{0pt}[0pt][0pt]{\elvrm t\sub{7}}}}
\put(187,686){\makebox(0,0)[b]{\raisebox{0pt}[0pt][0pt]{\elvrm t\sub{1}}}}
\put(187,640){\makebox(0,0)[b]{\raisebox{0pt}[0pt][0pt]{\elvrm t\sub{2}}}}
\put(187,593){\makebox(0,0)[b]{\raisebox{0pt}[0pt][0pt]{\elvrm t\sub{3}}}}
\put(187,548){\makebox(0,0)[b]{\raisebox{0pt}[0pt][0pt]{\elvrm t\sub{4}}}}
\put(187,502){\makebox(0,0)[b]{\raisebox{0pt}[0pt][0pt]{\elvrm t\sub{5}}}}
\put(205,456){\makebox(0,0)[b]{\raisebox{0pt}[0pt][0pt]{\elvrm t\sub{6}}}}
\put(241,731){\makebox(0,0)[b]{\raisebox{0pt}[0pt][0pt]{\elvrm I}}}
\put(268,709){\makebox(0,0)[b]{\raisebox{0pt}[0pt][0pt]{\elvrm V\sub{1}}}}
\put(296,686){\makebox(0,0)[b]{\raisebox{0pt}[0pt][0pt]{\elvrm I\sub{2}}}}
\put(323,662){\makebox(0,0)[b]{\raisebox{0pt}[0pt][0pt]{\elvrm V\sub{3}}}}
\put(350,640){\makebox(0,0)[b]{\raisebox{0pt}[0pt][0pt]{\elvrm I\sub{4}}}}
\put(378,617){\makebox(0,0)[b]{\raisebox{0pt}[0pt][0pt]{\elvrm V\sub{5}}}}
\put(405,593){\makebox(0,0)[b]{\raisebox{0pt}[0pt][0pt]{\elvrm I\sub{6}}}}
\put(223,709){\makebox(0,0)[b]{\raisebox{0pt}[0pt][0pt]{\elvrm I}}}
\put(223,686){\makebox(0,0)[b]{\raisebox{0pt}[0pt][0pt]{\elvit -past}}}
\put(250,662){\makebox(0,0)[b]{\raisebox{0pt}[0pt][0pt]{\elvit zag}}}
\put(278,640){\makebox(0,0)[b]{\raisebox{0pt}[0pt][0pt]{\elvit -inf}}}
\put(305,617){\makebox(0,0)[b]{\raisebox{0pt}[0pt][0pt]{\elvit helpen}}}
\put(432,524){\makebox(0,0)[b]{\raisebox{0pt}[0pt][0pt]{\elvit zwemmen}}}
\put(432,571){\makebox(0,0)[b]{\raisebox{0pt}[0pt][0pt]{\elvrm V\sub{7}}}}
\put(432,548){\makebox(0,0)[b]{\raisebox{0pt}[0pt][0pt]{\elvrm V\sub{7}}}}
\put(305,640){\makebox(0,0)[b]{\raisebox{0pt}[0pt][0pt]{\elvrm V\sub{3}}}}
\put(278,662){\makebox(0,0)[b]{\raisebox{0pt}[0pt][0pt]{\elvrm I\sub{2}}}}
\put(250,686){\makebox(0,0)[b]{\raisebox{0pt}[0pt][0pt]{\elvrm V\sub{1}}}}
\put(332,617){\makebox(0,0)[b]{\raisebox{0pt}[0pt][0pt]{\elvrm I\sub{4}}}}
\put(332,593){\makebox(0,0)[b]{\raisebox{0pt}[0pt][0pt]{\elvit -inf}}}
\put(359,593){\makebox(0,0)[b]{\raisebox{0pt}[0pt][0pt]{\elvrm V\sub{5}}}}
\put(387,571){\makebox(0,0)[b]{\raisebox{0pt}[0pt][0pt]{\elvrm I\sub{6}}}}
\put(387,548){\makebox(0,0)[b]{\raisebox{0pt}[0pt][0pt]{\elvit -inf}}}
\put(359,571){\makebox(0,0)[b]{\raisebox{0pt}[0pt][0pt]{\elvit helpen}}}
\end{picture}
\end{center}
\caption{Head-Raising in Dutch\label{fig.crossb}}
\end{figure}
A more or less typical account of head-raising in Dutch, for instance, is given
in Figure~\ref{fig.crossb}.  This is the type of movement presumed to be
responsible for the cross-serial dependencies that form the basis of Shieber's
claim that Swiss-German is non-context-free~\cite{shiebe85}.  Bresnan, et
al.,~\cite{BrKaPeZa82} have pointed out that analyses such as these form a
non-recognizable set.  Consequently, it cannot be possible to capture this
account within $\LKP$, and, in fact, the definition we give fails to license
these structures.  Examining why this is the case provides some insight into
the kinds of natural properties of linguistic structures that correspond to
increased language-theoretic complexity.

In order to rule out the possibility of ``forking'' chains---of some nodes
participating in the licensing of multiple gaps---we have required chains to be
maximal in the sense that they include every node that is related by link to
any node in the chain.  Consequently, we can license overlapping chains only if
they are distinguished in some way.  The account works for English because we
can 
classify chains in English into a bounded set of types in such a way that no
two chains of the same type ever cross.  (This fact depends to a great
extent on the minimality requirement in the antecedent-government relation.)
This property can be stated as a principle:
\begin{itemize}
\item[]{\em The number of chains which overlap at any single position in the
tree is bounded by a constant.}
\end{itemize}
Our approach to chains will work for any account of language that satisfies
this principle.  Once again, the linguistics literature provides arguments that
such bounds exist, at least in some cases.  As we have already noted, Manzini's
{\em Locality Theory}~\cite{manzini92} implies that no more than two
\xb{A}-chains ever overlap.  Stabler~\cite{stable94} makes the stronger claim
that such bounds exist for all linguistically relevant relationships in all
languages.  

Leaving aside the possibility that it may be possible to account for
cross-serial dependencies in Dutch in other ways, we can note that accounts
employing structures such as the one in Figure~\ref{fig.crossb} fail to meet
the bound on overlapping chains.  This is despite the fact that, if one orders
the movements bottom-up, each movement meets the strictest conceivable locality
constraint---each head moves to the closest possible position (often stated as
the {\em Head Movement Constraint}).  The problem is that, even if the
movements are ordered in this way, each movement carries the target
positions of the prior movements along with it.  Thus, in the final structure
all chains of 
head-movement overlap.  Given that the number of heads participating in these
structures is arbitrary, there can be no {\em a priori} bound on the number of
overlapping chains.  Note that in the example the two {\em helpen} chains
($[\f{V}_3,t_3]$ and $[\f{V}_5,t_5]$) are indistinguishable.  Any attempt to
form a chain including any of these nodes will be required to include all four
and the result will not be linearly ordered.

\section{Conclusion}\label{sec.concl}
In this paper we have 
introduced a kind of descriptive complexity result for the
strongly Context-Free Languages---a language is strongly context-free iff
the set of trees analyzing the syntax of its strings is definable in $\LKP$
(modulo a projection).  Using this result we have sketched a couple of language
complexity results relevant to GB, namely, that free-indexation cannot, in
general, be 
enforced by CFGs, and that a specific GB account of English licenses
a strongly context-free language.  The first of these results is not likely to
come as a surprise to the GB community.  The appropriateness of free-indexation
as a fundamental component in linguistic theories has been questioned in the
more recent GB literature on purely linguistic (rather than complexity
theoretic) grounds.

The second result is more surprising.  We don't expect it to extend to
the whole range of human languages, that is, to any theory of Universal
Grammar.  Shieber~\cite{shiebe85} and
Miller~\cite{miller91} (to cite two examples) give fairly strong evidence that
there are 
constructions that occur in human languages that are beyond the CFLs, and hence
not possible to capture in $\LKP$. 
As expected, our definitions fail for these
constructions.  The fact that the
definition works for English is a consequence of the fact that, in the account
of English we capture, it is 
possible to classify chains into finitely many categories in such a way that
no two 
chains from a given category ever overlap. 
  GB-style analyses of the constructions studied by Shieber and by
Miller include positions in which an unbounded number of chains can overlap.
Our definition is unable to identify any well-formed
chains including these positions; indeed, there is unlikely to be any way to
distinguish these chains without the 
equivalent of unbounded indices.

As it stands, this result speaks only of the particular account of English we
capture.  The fact that this is context-free says nothing about the nature of
human language faculty, since the principle it depends upon is unlikely to be a
principle of Universal Grammar.  It does, however, raise the prospect of wider
results.  Extensions of our descriptive
complexity result to larger 
language complexity classes
could provide formal restrictions on the principles employed by GB theories
that would be sufficient to provide non-trivial generative capacity results for
those theories
without losing the ability to capture the full range of human language.  With
such extended characterizations
one might establish upper bounds on the complexity of human language in
general. 
The possibility that such results might be obtainable is suggested by the fact
that we find numerous cases in which the issues arising in our studies for
definability reasons, and ultimately for language complexity reasons, have
parallels that arise in the GB literature 
motivated by more purely linguistic
concerns.  This suggests that the regularities of human languages that are the
focus of the linguistic studies are perhaps reflections of properties of the
human language faculty that can be characterized, at least to some extent, by
language complexity classes.


\end{document}